%% file: main.tex
\newcounter{notebox}

\documentclass{article}
\usepackage{goodfire/goodfire}

\usepackage{microtype}
\usepackage{hyperref}
\usepackage{url}
\usepackage{booktabs}
\usepackage{amsmath}
\usepackage{macros}
\usepackage{lineno}
\usepackage[color=pink!20,textsize=scriptsize]{todonotes}

\usepackage{tcolorbox}

\definecolor{notebg}{RGB}{242,244,243}  % light gray fill
\definecolor{notefg}{RGB}{47,93,77}     % dark green text

\hypersetup{colorlinks=true, citecolor=primary, linkcolor=primary, urlcolor=primary}
\AtBeginDocument{%
  \hypersetup{colorlinks=true, citecolor=primary, linkcolor=primary, urlcolor=primary}%
}

\input{authors}
\author{\paperauthors}

\title{A Phenomenology of Neural Geometry in Vision-Language-Action Models}

\begin{document}
\vspace*{-5.5mm}

\maketitle

\begin{figure}[h]
    \vspace{-5mm}
    \centering
    \hspace*{-0.05\textwidth}\includegraphics[width=1.1\textwidth]{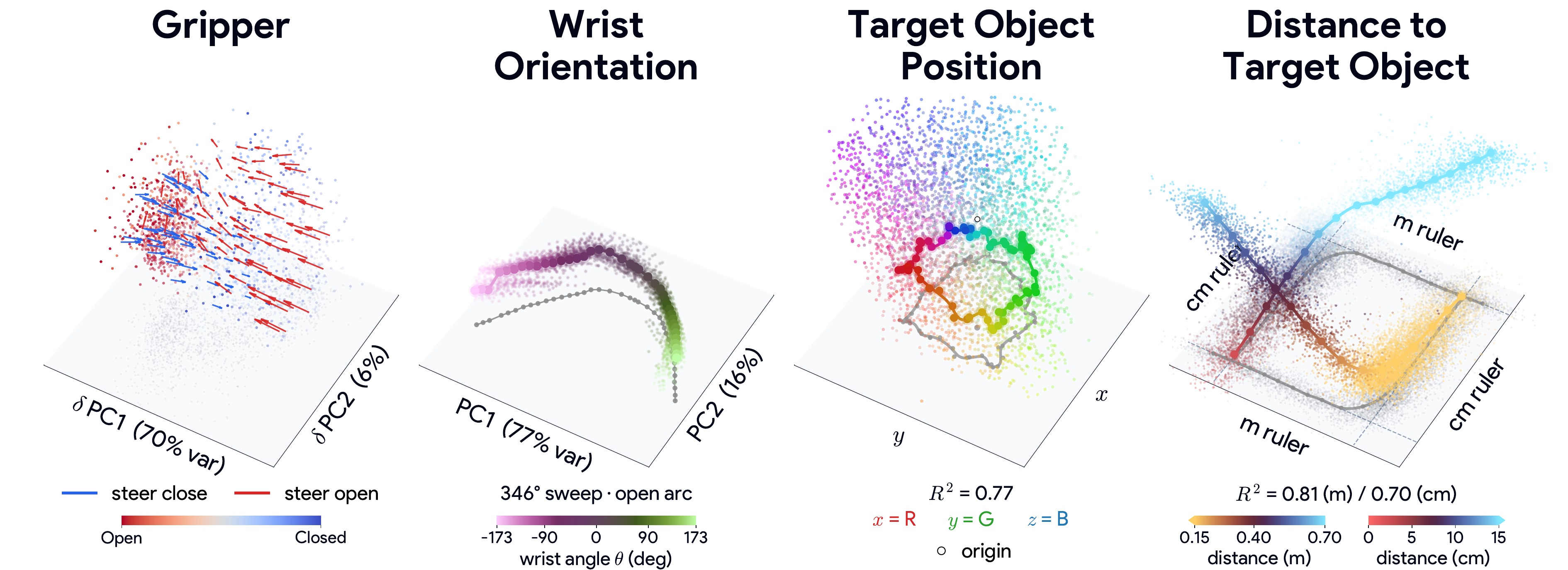}
    \vspace{-5mm}
    \caption{{\small\textbf{Control variables realize distinct geometries in a VLA policy.} MolmoBot~\citep{deshpande2026molmob0t} activations over many states (clouds) and controlled sweeps varying one physical quantity (traces). Clouds and traces are activations projected onto a low-dimensional basis (principal components for the gripper and wrist; the span of a linear probe's weights for target position); for distance, the activations along the two ruler directions are plotted against the true distance (z axis). (\textbf{Gripper}) Open and closed states from natural rollouts separate along one axis; arrows show interventions reversing the command. (\textbf{Wrist}) Rotation sweeps pooled over 64 houses trace an open arc, not the circle unconstrained rotation would give. (\textbf{Target position}) A structured trajectory within the cloud of teleported target placements. (\textbf{Distance}) Two orthogonal rulers, coarse for approach and fine for grasp.}}
    \vspace{-3mm}
    \label{fig:headline}
\end{figure}

\vspace{-0mm}

\begin{abstract}
%%%%
Robotic manipulation is inherently geometric, but what shape does the world take inside the policy that performs it?
Prior work on vision-language-action (VLA) policies establishes that task-relevant variables are linearly readable from activations, yet readability certifies only that a variable is present and says nothing about the shape its values trace, which may be curved and multi-dimensional.
Using controlled sweeps, natural rollouts, linear probes, and activation interventions, we build a taxonomy of seven task-relevant concepts in VLA models.
We find that realized geometry is predominantly multi-dimensional and is shaped by the constraints under which the policy acts. Physical joint limits, for example, leave wrist orientation an open arc rather than a circle, and distance to the target, a single scalar, is realized in two orthogonal subspaces, a coarse ruler for the approach and a fine ruler for the grasp.
The same geometries reappear in two further VLA models on real-world robot data, indicating that they are neither model-specific nor artifacts of simulation.
We demonstrate how the findings are actionable by steering along the geometry a concept realizes.
Overall, this work lays out an initial phenomenology of realized geometry in VLA policies and demonstrates that a concept's shape carries as much importance as its readability.
\end{abstract}

\vspace{-2mm}
\section{Introduction}
\vspace{-2mm}

Robotic manipulation is inherently geometric. Robotic systems operate in three-dimensional space through articulated kinematic structures, while the objects they manipulate and the tasks they perform have their own geometric organization, including symmetries, constraints, and phases of interaction. Any general manipulation policy must contend with this structure, and vision-language-action (VLA) models have become the dominant paradigm for building such policies \citep{brohan2023rt, kim2024openvla, team2024octo, black2024pi_0, intelligence2025pi_, lee2025molmoact, fang2026molmoact2, deshpande2026molmob0t}.
VLAs inherit visual and semantic priors from backbones pretrained on internet-scale data, including vision encoders \citep{radford2021learning, oquab2023dinov2, zhai2023sigmoid}, language models \citep{touvron2023llama, team2024gemma}, and full vision-language models \citep{driess2023palm, beyer2024paligemma, deitke2025molmo, wang2024qwen2}.
This data has grown by orders of magnitude in volume and diversity \citep{o2024open, walke2023bridgedata, bu2025agibot, deshpande2026molmob0t, bjorck2025gr00t}, with generalization improving along power-law-like trends \citep{lin2025data}.
The resulting systems are increasingly capable, general, and even deployed in the real world \citep{intelligence2025pi_, intelligence2025pi, team2025gemini, figure2026helix02, gemini2026robotics2, figure2025helix, figure2025bmw}.

How such competence is reflected in the representations of these VLAs is the subject of a nascent literature~\citep{qi2025control, qi2026action, swann2026sparse, buurmeijer2026observing, haon2025mechanistic, bhardwaj2026decoding, lu2025probing, molinari2025emergent, khan2025controlling, grant2026not, goyal2026hitchhiker, yang2026interpretable, shi2026vla}. 
In particular, one line of work uses probes, testing whether a quantity of interest can be read out from the model's activations \citep{joseph2026interpreting, buurmeijer2026observing, bhardwaj2026decoding, lu2025probing, molinari2025emergent, grant2026not, yang2026interpretable}. 
A second trains sparse autoencoders (SAEs)~\citep{cunningham2023sparse, bussmann2024batchtopk, rajamanoharan2024jumping, fel2025archetypal, costa2025flat, hindupur2025projecting, gao2025scaling, bricken2023monosemanticity, lubana2026priors, bhalla2026temporal, templeton2024scaling, rajamanoharan2024improving, lieberum2024gemma, thasarathan2025universal, makhzani2013k} to discover concepts without supervision \citep{swann2026sparse, khan2025controlling, grant2026not}, and the discovered features even provide control over the policy via steering~\citep{joseph2025steering}.

Both lines of work share a limitation. A probe can ensure that a variable is decodable, yet says nothing about the structure through which the model ``realizes'' it, and an SAE inherits the linear representation hypothesis (LRH)~\citep{park2023linear, zheng2026model}, which casts each concept as a single direction in activation space and therefore fixes that structure in advance.
Interpretability for language and vision models, conscious of this limitation, has moved beyond the single direction to multi-dimensional features and curved geometry~\citep{modell2025origins,feucht2026arithmetic,wurgaft2026manifold,bhalla2026sparse, fel2026into, fel2026structuring, engels2025not, lubana2026priors, karkada2026symmetry, dooms2025finding, kantamneni2025language, gurnee2026models, yocum2025neural, park2025iclr, sarfati2026shape, costa2025flat, pearce2025tree, diego2024polar, chung2018classification, cohen2020separability, hosseini2026context, dhimoila2026cross, fel2025archetypal, gorton2024missing, sun2025dense, tvetkova2025convex, park2025geometry, park2026information}, a line of work now gathered under the name of \emph{Neural Geometry} \citep{geiger2026world}.

Do VLAs realize their variables through similarly rich geometry? Two facts suggest they might.
First, vision models, including the backbones VLAs inherit, already carry colors, shadows, and position as multi-dimensional features rather than single directions~\citep{fel2026into, fel2026structuring, pach2026latent, uselis2026compositional}.
Second, representations tend to inherit the symmetries of their training data, so that cyclic concepts, for example, come to be arranged on circles~\citep{karkada2026symmetry}.
The variables a manipulation policy must track, such as poses, distances, contacts, and task phase, are geometric objects with symmetries of their own before the network ever represents them.

We take up this question and make the following contributions:

\begin{itemize}[nosep,leftmargin=*]
    \item \textbf{Realized geometry as a lens for VLA interpretability.} We formalize the distinction between \emph{concept realization}---the capacity to extract information about a concept under a given readout---and its \emph{realized geometry}---the structure activations trace as the concept varies. Linear readability establishes only that a concept is present, while its realized geometry determines how its values are structured and whether they can be manipulated.
    \item \textbf{A taxonomy of realized geometry in MolmoBot.} Combining controlled sweeps, natural rollouts, linear probes, and activation interventions, we chart the realized geometry of seven control concepts: gripper state, end-effector position, distance to the target, target object position, contact, task phase, and wrist orientation. We find this geometry is predominantly multi-dimensional and shaped by the constraints under which the policy acts: joint limits leave wrist orientation an open arc rather than a circle, while the numerical precision of the embedding splits distance into orthogonal coarse (meter-scale) and fine (centimeter-scale) rulers.
    \item \textbf{Real-world transfer.} We reproduce these geometries in $\pi_{0.5}$ \citep{intelligence2025pi_} and MolmoAct2 \citep{fang2026molmoact2} on the real-world DROID dataset \citep{khazatsky2024droid}, indicating they are neither model-specific nor artifacts of simulation.
\end{itemize}

\vspace{-2mm}
\section{Realization of Control Geometry}
\label{sec:theory}
\vspace{-2mm}

\input{realization}

\begin{figure}[tb]
    \centering
    \includegraphics[width=1.0\linewidth]{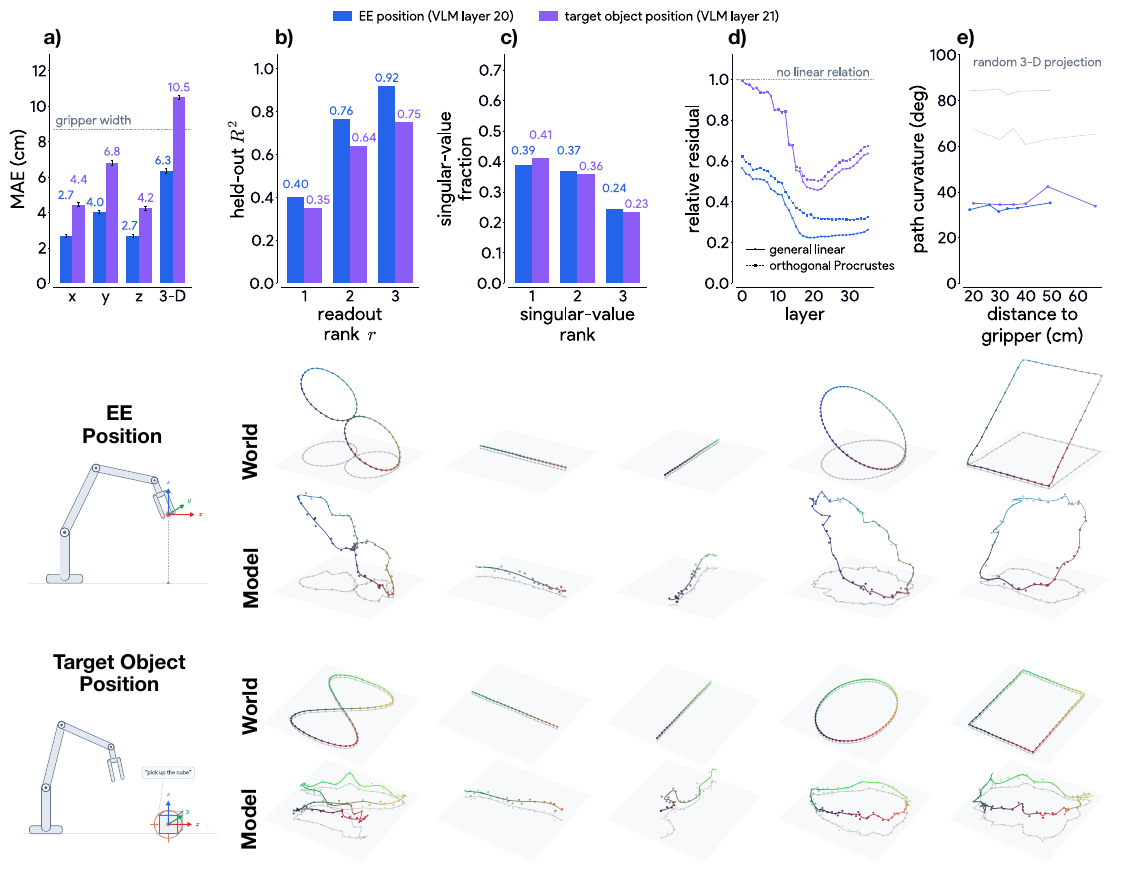}
    \caption{\textbf{MolmoBot's VLM encodes its end-effector (layer 20) and the target object (layer 21) in separate, approximately Euclidean 3-D subspaces.}
    \textbf{Top:} (a) held-out per-axis error relative to gripper width; (b) held-out $R^2$ vs.\ readout rank; (c) singular values of the rank-3 readout; (d) residual of a linear map from world to activation displacements across layers (solid: general linear; dashed: orthogonal; reference line: no linear relation); (e) curvature of straight world paths inside each subspace vs.\ distance to the gripper, compared with a random 3-D projection.
    \textbf{Middle, bottom:} five gripper paths through the workspace (middle) and five target paths across its support surface (bottom), shown in world coordinates (upper) and projected into the corresponding subspace (lower). Subspaces are fit with the tested house held out.
    }
    \label{fig:position}
\end{figure}
\vspace{-4mm}

\begin{figure}[tb]
    \centering
    \includegraphics[width=\linewidth]{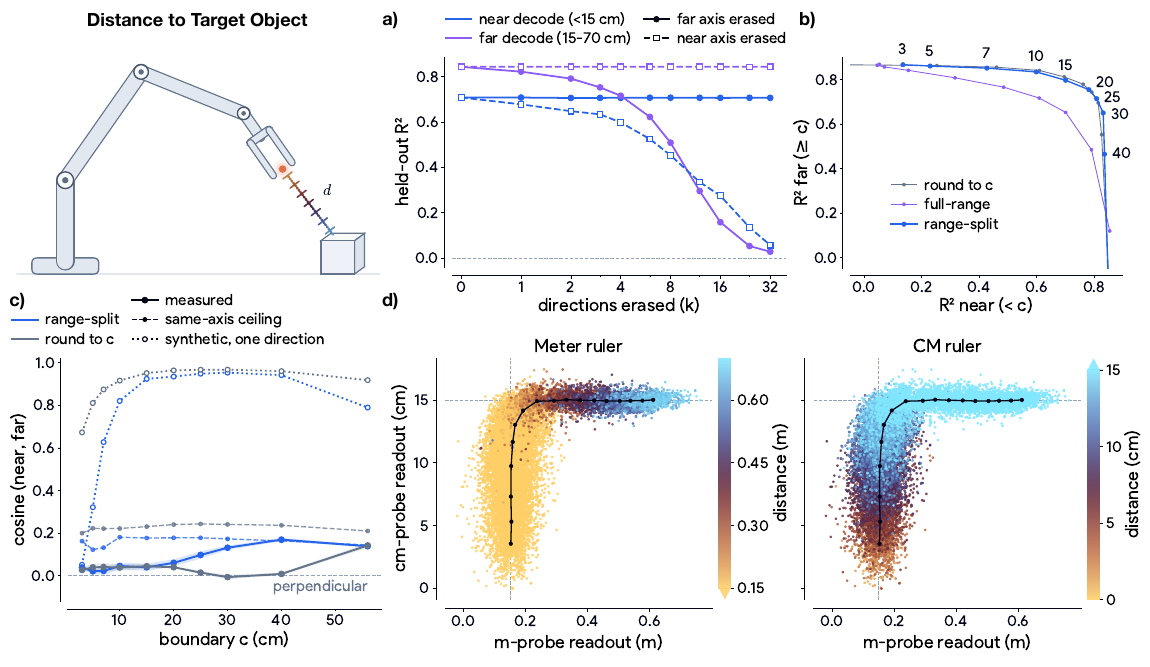}
    \caption{\textbf{MolmoBot's VLM encodes distance to the target in nearly orthogonal near- and far-range codes (layer 21).}
    (a) Near (${<}15$\,cm) and far (15--70\,cm) probe $R^2$ after erasing the top-$k$ directions of the other probe, computed against each range's own variance. Erasing far directions leaves near decoding intact while far decoding collapses, and vice versa. The intact ($k=0$) probes have errors of ${\sim}1.3$\,cm (near) and ${\sim}3.6$\,cm (far).
    (b) Near $R^2$ (${<}c$) vs.\ far $R^2$ (${\geq}c$) as the cutoff $c$ is swept. Clipped (`round to c') probes match probes fit separately on each range and outperform a single full-range probe.
    (c) Cosine between the near- and far-probe weight vectors as the cutoff $c$ is swept, for range-split and clipped (`round to c') pairs (fold mean $\pm$ s.d.). For $c \lesssim 30$\,cm, both cosines stay below the cosine between two copies of the same probe fit on different halves of the data (dashed). On synthetic activations that encode distance along a single direction, the same probes are nearly parallel for $c \gtrsim 15$\,cm (dotted).
    (d) Activations plotted by their meter- and centimeter-probe readouts, colored by true distance in meters (left) and centimeters (right). Far samples vary along the meter axis while the centimeter readout saturates, and near samples do the opposite, tracing an L.}
    \label{fig:dist_target}
\end{figure}

\section{A Taxonomy of Realized Geometry in MolmoBot}
\label{sec:molmo}
We now build a taxonomy of how a model realizes the variables of pick and pick-and-place tasks. Our subject is MolmoBot \citep{deshpande2026molmob0t}, which pairs a VLM that reads the language instruction and camera images of the scene with an action expert that reads the robot's proprioceptive state and generates action chunks via flow matching, cross-attending to the VLM at each layer (Appendix~\ref{app:architecture}). We choose MolmoBot for three reasons: it is a relatively strong policy on numerous benchmarks \citep{deshpande2026molmob0t}, and we hypothesize that stronger models realize task geometry more faithfully; its weights are open, so we can record and edit activations; and it runs in MolmoSpaces \citep{kim2026molmospaces}, which provides ground-truth task variables and lets us intervene on the world mid-rollout. We study seven variables in three groups: spatial geometry (EE position, target object position, and distance to the target), task structure (contact and task phase), and action geometry (wrist orientation and gripper aperture). For each, we ask whether the model represents it and with what geometry.

\subsection{Spatial Geometry}
We begin with spatial geometry in the VLM backbone, which receives no proprioception and must infer these variables from images: EE position, target object position, and distance to the target.

\paragraph{End-effector and target object position.}
Since both positions live in $\mathbb{R}^{3}$, we expect a faithful code for each to occupy a three-dimensional subspace. We teleport the arm through poses above the workspace and the target through a uniform sweep of positions on the workspace surface, record VLM activations after each teleport, and fit ridge probes onto $(x,y,z)$, evaluated on held-out houses (Appendix~\ref{app:position}). For both variables, three dimensions are necessary and sufficient: held-out $R^2$ rises sharply up to rank 3 and does not improve beyond it, reaching $0.92$ for EE position (layer 20) and $0.75$ for the target (layer 21), with per-axis error within the gripper's width (Fig.~\ref{fig:position}a--c). These subspaces also preserve spatial geometry: translating the gripper or target by a known world vector shifts the activations in a way well fit by an orthogonal (Procrustes) map (Fig.~\ref{fig:position}d), and straight world paths stay nearly straight inside each subspace (Fig.~\ref{fig:position}e). Position thus sets the taxonomy's baseline: from pixels alone, the VLM re-encodes a Euclidean quantity at its native dimension and approximately its native metric.\footnote{The action expert also carries rank-3 subspaces for both variables (Appendix Figs.~\ref{fig:position_ae} and~\ref{fig:target_pos_ae}).} We next turn to a variable that couples the two: distance to the target.

\paragraph{Distance to target object.}
Distance to the target is a scalar, but the policy needs it at two very different scales: it must tolerate noise during the approach yet resolve the grasp to within a centimeter. A single direction with finite precision may struggle to carry both, so we hypothesize that the model encodes distance in more than one subspace. We fit separate linear probes on near (${<}15$\,cm) and far (15--70\,cm) states and iteratively erase one probe's directions from the activations (Appendix~\ref{app:distance}). Each code survives removal of the other: erasing 32 far directions drops far $R^2$ from $0.84$ to near zero while near $R^2$ stays at $0.71$, and vice versa (Fig.~\ref{fig:dist_target}a).

To isolate the two codes, we fit a meter probe with targets clipped below 15\,cm and a centimeter probe with targets clipped above 15\,cm, a cutoff that is Pareto-optimal for both ranges (Fig.~\ref{fig:dist_target}b). Each probe reads its own range well ($R^2 = 0.81$ and $0.70$) and saturates outside it, so activations projected onto the two readouts trace an L (Fig.~\ref{fig:dist_target}d). The two probe directions are nearly orthogonal, with a cosine of $0.04$ against a split-half ceiling of ${\sim}0.23$ (Fig.~\ref{fig:dist_target}c). Clipping does not impose this orthogonality, since a probe can only select directions the activations contain. On synthetic activations that encode distance along a single direction, the near and far probes at the 15 cm boundary come out nearly parallel (cosine $0.95$ for the clipped pair, $0.92$ for the range-split pair), so neither clipping nor range-restricted fitting yields orthogonal probes on its own. In the model, probes fit separately on each range, without clipping, are also nearly orthogonal ($0.04$ vs.\ ${\sim}0.18$; Fig.~\ref{fig:dist_target}c). Distance thus departs from the position baseline: the model realizes a single scalar along separate directions for coarse and fine scales, consistent with the limited precision of a representation bending how a concept is encoded. 

\begin{figure}[tb]
    \centering
    \includegraphics[width=\linewidth]{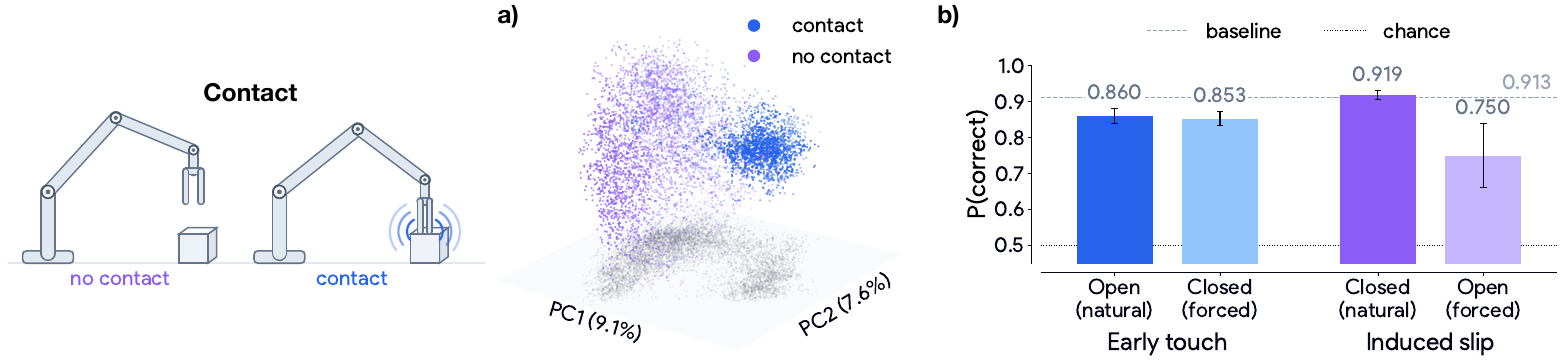}
    \caption{\textbf{MolmoBot's VLM encodes contact with the target along a direction that survives off-schedule contact (layer 24).}
    (a) Natural rollouts projected onto their first three principal components, colored by contact; the two states separate cleanly.
    (b) An LDA probe trained on natural rollouts decodes contact with 91.3\% accuracy. When we force an early touch or an induced slip by teleporting the object, the same probe still decodes contact well above chance.
    }
    \label{fig:contact}
\end{figure}

\begin{figure}[tb]
    \centering
    \includegraphics[width=\linewidth]{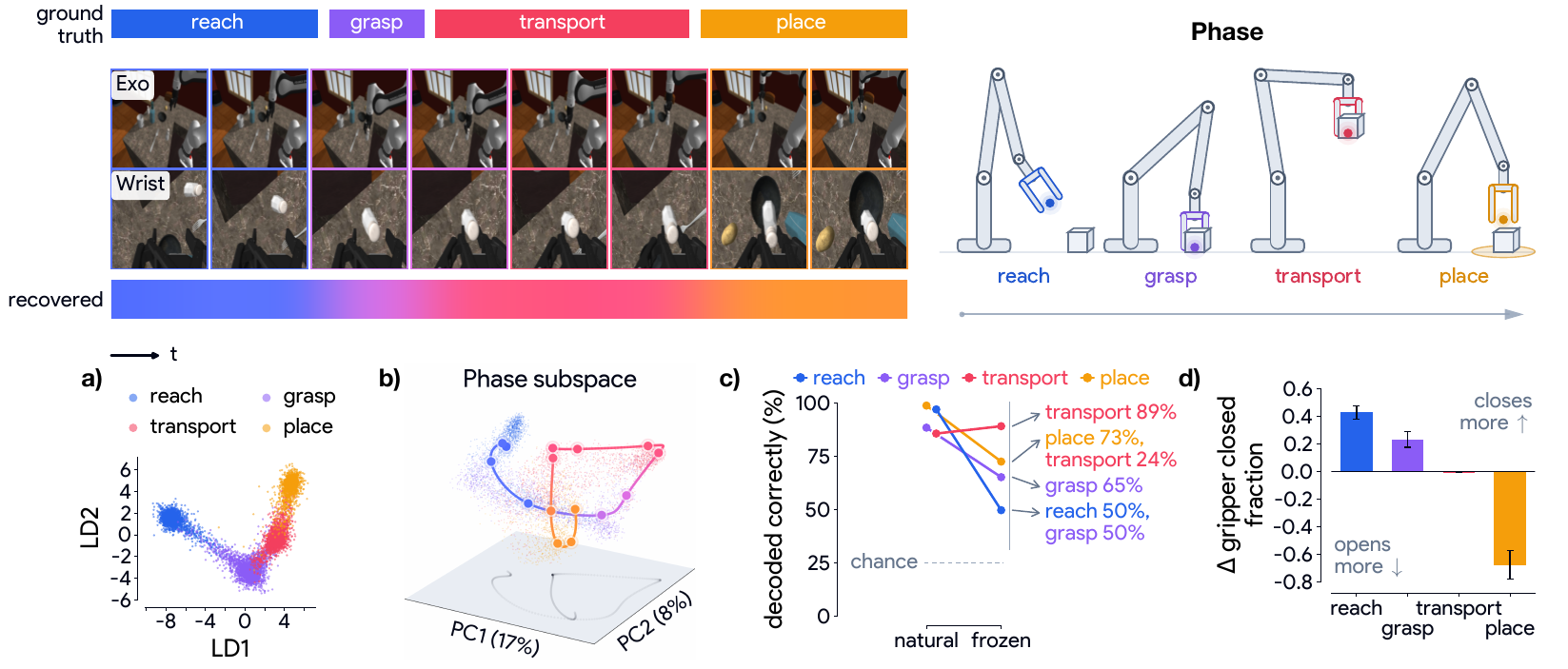}
    \caption{\textbf{MolmoBot's action expert encodes task phase, but infers it from motion, so pauses are read as phase transitions (layer 24).}
    \textbf{Top:} an example pick-and-place trajectory and its path through the phase subspace.
    \textbf{Bottom:} 
    (a) LDA projection of natural rollouts, colored by phase. 
    (b) First three principal components of natural activations, colored by true phase; the example trajectory is colored by the LDA probe readouts.
    (c) Freezing the arm shifts the decoded phase toward grasp during reach; frozen place landmarks are most often still decoded as place. Right: most frequent decoded phases when frozen (top two for reach and place).
    (d) Change in the fraction of executed-chunk steps commanding a closed gripper (frozen $-$ natural; mean $\pm$ 95\% CI across landmarks). Frozen reach and grasp moments close the gripper, while frozen place moments open it (Fig.~\ref{fig:phase-freeze}).
    }
    \label{fig:phase}
\end{figure}

\subsection{Task Structure}
We next turn to variables that describe the structure of the task rather than the geometry of the scene: contact with the target object, in the VLM, and task phase, in the action expert.

\paragraph{Contact.}
Contact with the target is binary, so we treat it as a two-class problem and use linear discriminant analysis (LDA) to find the direction that separates the classes. We label each state of the rollouts by simulator contact and fit the discriminant on VLM activations (layer 24). Contact and no-contact states separate cleanly even in an unsupervised PCA of natural activations (Fig.~\ref{fig:contact}a), and the discriminant decodes contact with 91.3\% accuracy. Because contact co-occurs with other variables in natural rollouts, high accuracy alone does not show that the direction tracks contact itself. We therefore force contact off schedule by teleporting the object into the gripper early (an early touch) or away from it during a grasp (an induced slip). Even though both interventions are out of distribution for the model, the same direction still decodes contact well above chance (Fig.~\ref{fig:contact}b), indicating that the model maintains a real contact code rather than some proxy for task timing.

\begin{tcolorbox}[colback=notebg, colframe=notebg, arc=4mm, float=tb,
                  boxsep=0pt, top=4pt, bottom=4pt, left=6pt, right=6pt]
\color{notefg}
\refstepcounter{notebox}\label{box:belief}%
\textbf{Box~\thenotebox: Belief vs.\ Action.} Probe accuracy does not distinguish the two regimes.
\smallskip

\noindent
\textbf{Belief.}~Some structures encode the model's estimate of its current state, such as wrist angle, or its relationship to the scene, such as the distance to the object. Editing such structures effectively amounts to affecting the policy's perception and can thus only affect behavior indirectly. The policy always seeks to close the gap between perceived state and goal state, meaning the behavior will shift against the edit, producing what we call a \emph{compensation effect}.

\smallskip
{\color{notefg!35}\hrule height 0.2pt}
\smallskip

\noindent
\textbf{Action.}~Other structures directly represent the commanded behavior, such as the gripper. Editing such structures affects behavior directly, while leaving other outputs, such as other joints, unchanged. To identify these edits, we reverse engineer the desired behavior by pulling a target output action back into activation space.

\smallskip
\noindent
\end{tcolorbox}

\paragraph{Phase.}
We now move to the action expert, where we expect variables coupled to actions to appear most strongly. Unlike contact, task phase is defined by the whole trajectory rather than a single state, so we fit a multi-class LDA on activations along full pick-and-place trajectories (AE, layer 24; labels in Appendix~\ref{app:phase}). Natural activations separate into distinct clusters for reach, grasp, transport, and place, both in the LDA projection and in the first three principal components (Fig.~\ref{fig:phase}a--b). 

To test how the model infers phase, we freeze the arm for several actions. Freezing during reach shifts the decoded phase toward grasp, while freezing during grasp, transport or place mostly leaves it unchanged (Fig.~\ref{fig:phase}c). The shifts are also behavioral: freezing raises the fraction of steps commanding a closed gripper by 43 percentage points during reach and lowers it by 68 points during place, while leaving it unchanged during transport (Fig.~\ref{fig:phase}d).

MolmoBot thus represents contact and task phase with different robustness. While contact remains decodable when forced off schedule, phase does not reliably survive a pause. Since the model does not receive previous actions, it cannot keep an explicit time index and must re-derive phase from motion at each chunk. We hypothesize that in demonstrations, the arm stops when the gripper actuates, so that a still arm is read as a phase transition and the model issues the corresponding gripper command.

\subsection{Action Geometry} % Behavior Geometry
So far, we have studied variables by reading them out. For the last two, wrist orientation and gripper aperture, we also write to them, which reveals a distinction that readout alone cannot: whether an edit lands on what the model believes or on what it commands (Box~\ref{box:belief}).

\begin{figure}[tb]
    \centering
    \includegraphics[width=\linewidth]{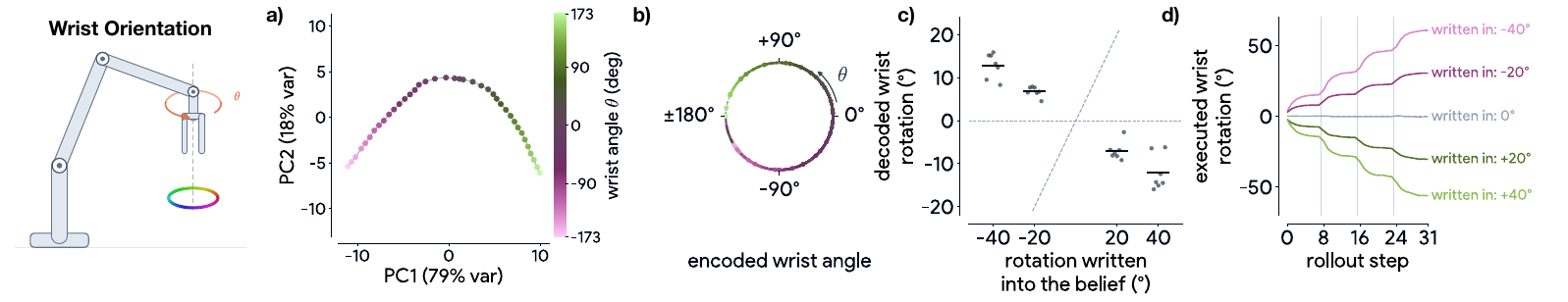}
    \caption{\textbf{The action expert encodes wrist orientation as an open arc, and edits to it act as beliefs (layer 9).}
    We sweep the wrist's roll joint through its full range ($-172.8^\circ$ to $+172.8^\circ$), holding other joints fixed, and record activations at each angle (camera views in Fig.~\ref{fig:wrist_frames}).
    (a) The sweep traces an open curve: its endpoints, nearly the same physical pose, lie far apart.
    (b) An autoencoder maps activations onto an exact circle.
    (c) A single $\pm 20^\circ$ or $\pm 40^\circ$ edit produces a decoded rotation opposing the written angle (dots: seven held-out scenes; ticks: means; dashed diagonal: the response a command would produce).
    (d) Holding the edit at every step compounds the corrections, rotating the wrist opposite the written angle by about its magnitude (vertical lines mark action chunks).}
    \label{fig:wrist}
\end{figure}

\begin{figure}[t]
    \centering
    \includegraphics[width=\linewidth]{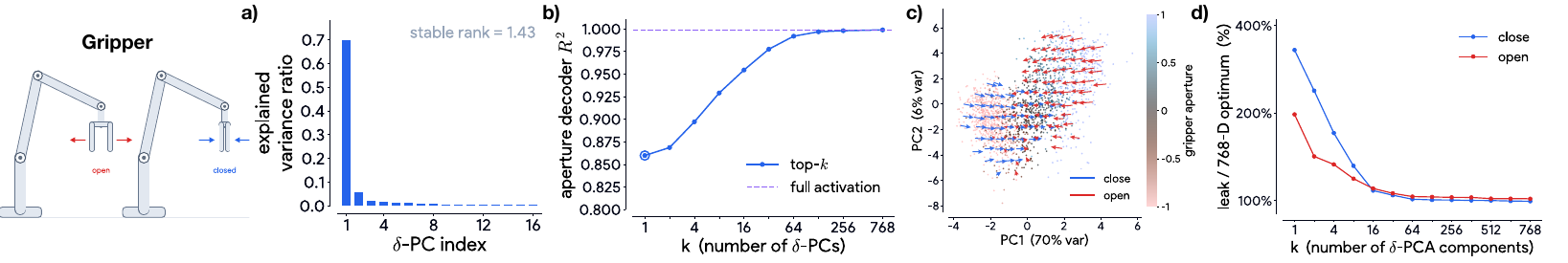}
    \caption{
    \textbf{Gripper control is one-dimensional (AE, layer 32).}
    (a) The first principal component of the pullback steering vectors captures ${\sim}70\%$ of their variance.
    (b) One principal component of the pullback steering vectors decodes the commanded gripper aperture nearly as well as the full activation.
    (c) Natural-rollout activations colored by gripper state; the steering vectors point from each cluster toward the other.
    (d) One component flips the gripper but leaks into the arm joints relative to using the full 768-dimensional steering vector; more components remove this excess leakage.
    }
    \label{fig:gripper}
\end{figure}
\vspace{-3mm}

\paragraph{Wrist orientation.}
Orientation is cyclic, so its most compact code would be a circle. However, the wrist's roll joint is limited to $\pm 172.8^\circ$ and never completes a full turn, and the action expert receives this angle directly as a raw, non-wrapping joint reading. Consistent with this, sweeping the joint through its range (Appendix Fig.~\ref{fig:wrist_frames}) traces an open arc in the action expert (layer 9), whose ends lie far apart in activation space despite being only $14^\circ$ apart physically (Fig.~\ref{fig:wrist}a). The action expert thus encodes an unwrapped angle: after a full turn past the joint limit (overridden in simulation), the images are identical but its activation does not return to the start, while the VLM's does (Appendix Fig.~\ref{fig:wrist_multiturn}). To write to this code, we cannot add one fixed vector: on an open arc, the direction that advances the rotation depends on where the activation sits (Fig.~\ref{fig:wrist}a). We instead train a small MLP autoencoder that maps activations onto an exact circle (Fig.~\ref{fig:wrist}b), so that rotating by $\theta$ is the same operation at every state; the edit is the decoded difference between the rotated and unrotated points (Appendix~\ref{app:wrist}). This edit is built from the representation alone and does not command motion, making it a belief edit (Box~\ref{box:belief}). Applied once mid-rollout, it produces a compensation effect: the decoded action rotates $0.33 \pm 0.08$ against each written degree (Fig.~\ref{fig:wrist}c), as if the policy were closing the gap between its believed angle and its goal. Held at every query, the corrections compound, and the wrist rotates opposite the edit by $1.0$--$1.3\times$ the written angle (Fig.~\ref{fig:wrist}d). Thus, even a geometry-matched edit lands on a belief, so for the gripper we work from the action side.

\begin{figure}[H]
    \centering
    \includegraphics[width=\linewidth]{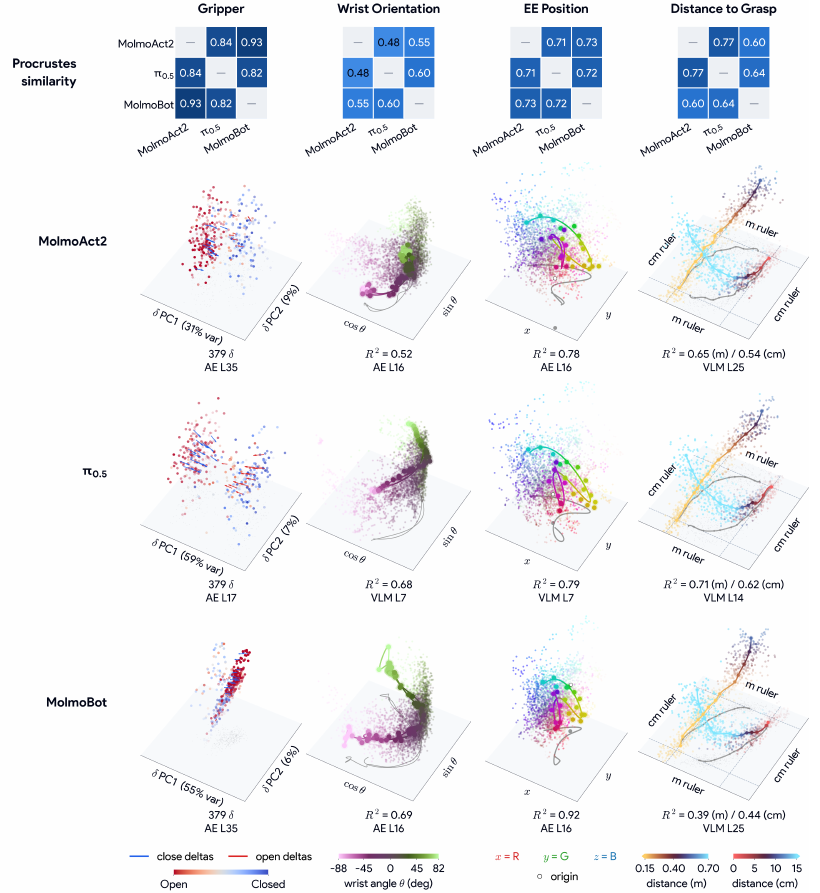}
    \caption{\textbf{The geometry recovered in MolmoBot largely reappears in $\pi_{0.5}$ and MolmoAct2 on real-world DROID data.}
    Procrustes similarity between each pair of models is highest for the gripper and lowest for wrist orientation, and all values exceed a shuffled null ($0.00$--$0.02$ for gripper and distance; $0.11$--$0.18$ for wrist and EE position).
    (\textbf{Gripper}) Pullback steering vectors (379 per model) on their top three principal components, colored by whether they close (blue) or open (red) the gripper.
    (\textbf{Wrist orientation}) Activations on a $(\cos\theta,\sin\theta)$ probe, colored by angle.
    (\textbf{EE position}) Activations on the $(x,y,z)$ probe.
    (\textbf{Distance to grasp}) The real-world proxy for distance to target: activations on the meter and centimeter rulers of Fig.~\ref{fig:dist_target}, with far (cool) and near (warm) points spreading along different rulers.
    Episode trajectories are overlaid in the last three columns.}
    \label{fig:real_manifolds}
\end{figure}

\paragraph{Gripper.}
The gripper command is an action variable (see Box~\ref{box:belief}), so a natural question is, then, how can we edit just the gripper action? Here, we derive the edit from the action output. For each sampled rollout state, we use pullback \citep{wurgaft2026manifold} to find a steering vector $\bm{\delta}$, added as $\bm{h} \leftarrow \bm{h} + \bm{\delta}$ in the action expert (layer 32), optimized with gradients from the action output until the decoded gripper command flips while the arm joints stay fixed. Natural activations form two clouds by gripper state, and each successful steering vector points from its own cloud toward the other (Fig.~\ref{fig:gripper}c), so edits derived only from action gradients recover the axis that separates the states. These edits are largely one-dimensional: their first principal component captures ${\sim}70\%$ of their variance (Fig.~\ref{fig:gripper}a) and is sufficient to accurately decode the commanded gripper state (Fig. \ref{fig:gripper}b). This direction is not fully isolated, however: steering along it also moves the arm joints, and suppressing this leakage requires additional components (Fig.~\ref{fig:gripper}d).

Together, the wrist and gripper show two things. First, physical limits shape realized geometry: a variable we hypothesized would be circular is encoded as an arc. Second, recovering a variable's geometry does not determine how editing it affects behavior: edits along the wrist's arc land on a belief and elicit compensation, whereas edits derived from gripper action gradients change the action.

\section{Representations in Real Robots}\label{sec:real-robots}

All experiments in Section~\ref{sec:molmo} were run in simulation. Since the gap between simulation and reality remains a central open problem in robotics \citep{jakobi1995noise, tobin2017domain, kadian2020sim2real}, we ask whether the geometry we found carries over to real data. We repeat our analyses on MolmoBot and two additional VLAs, $\pi_{0.5}$ \citep{intelligence2025pi_} and MolmoAct2 \citep{fang2026molmoact2}, using episodes from the real-world DROID dataset \citep{khazatsky2024droid}. Real data lacks the privileged information of simulation, such as ground-truth object poses and mid-rollout interventions, so we focus on four variables we can measure: gripper aperture, wrist orientation, EE position, and distance to target, for which we use distance to the eventual grasp location as a proxy.

All four variables are recovered in every model, with cross-model Procrustes similarities well above a shuffled-variable null (Fig.~\ref{fig:real_manifolds}). The gripper structure is the most consistent across models: pullback steering vectors again separate closing from opening, with the first principal component carrying $31$--$59\%$ of their variance. EE position is linearly decodable in all three models ($R^{2}$ of $0.78$--$0.92$), and the probed activations follow the arm smoothly through an episode. Wrist orientation is decoded less well ($R^{2}$ of $0.52$--$0.69$) and agrees least across models. Most notably, the two-scale distance code reappears: in all three models, far points spread along the meter ruler while near points split off onto the centimeter ruler. This structure is sharper in $\pi_{0.5}$ and MolmoAct2 than in MolmoBot, consistent with MolmoBot being evaluated on real data outside its simulated training distribution. Together, these results suggest that the geometry we identified is neither an artifact of simulation nor specific to one model, but a recurring way in which VLAs organize the variables needed for manipulation.

\section{Conclusion}
\label{sec:conclusion}
In this work, we asked what geometry manipulation variables realize inside VLAs. In MolmoBot, the answer is rarely a single direction: position occupies a near-Euclidean 3-D subspace, wrist orientation an open arc imposed by joint limits, and distance two nearly orthogonal rulers, and this structure largely recurs in $\pi_{0.5}$ and MolmoAct2 on real data. Two lessons follow. First, decodability is only a starting point: a variable can be linearly readable yet require a geometry-matched edit to steer, and even then the edit may land on a belief the policy compensates against rather than an action it executes. Interpretability tools for VLAs should therefore match both the geometry a concept realizes and its role in control. Second, this geometry is shaped by the policy's inputs and physical limits, so design choices such as how joint angles are encoded shape how the model represents them. Our taxonomy is supervised, but it provides ground truth against which unsupervised methods, such as block-sparse featurizers~\citep{fel2026structuring}, can be validated. We believe the geometry of VLAs is a promising direction for both interpretability and better policies.

\subsubsection*{Acknowledgments}
We thank our colleagues at Goodfire for valuable feedback at various stages of the project.

\bibliography{main}
\bibliographystyle{goodfire/bibliostyle}

\newpage
\appendix

\section{Realization does not determine Geometric Realization.}
\label{app:relization_vs_geometry}
For example, linear readability does not imply linear realized geometry. Consider the map
\begin{equation}
    \realmap\left(t\right)
    =
    \begin{pmatrix}
        t \\
        \cos t \\
        \sin t
    \end{pmatrix}.
    \label{eq:helix}
\end{equation}
The value \(t\) is exactly linearly readable through
\(\gamma\left(\bm{a}\right)=a_1\), yet its realized geometry is a helix. The concept is intrinsically one-dimensional, but its trace is curved and occupies several activation dimensions. Thus, the subspace used by a linear readout should not be confused with the shape traced by the concept itself.

This distinction also matters for control. A single fixed steering direction \(\bm{v}\) can follow a concept globally only when
\begin{equation}
    \realmap_{i,\bm{c}}\left(\bm{u}+\bm{\delta}\right)
    -
    \realmap_{i,\bm{c}}\left(\bm{u}\right)
    \approx
    \bm{v}\bm{\delta}
    \label{eq:fixed-steering}
\end{equation}
throughout the relevant range. For a curved or multiscale geometry, the appropriate activation steering vector instead changes with the current concept value. 
Another way to put it is that a readout specifies how to measure a concept from an activation but it does not necessarily specify how to move the activation while remaining on the geometry realized by that concept. 
Linear readability therefore does not imply linear steerability.

\input{extended_results}

\clearpage
\input{experimental_details}

\end{document}

%% file: authors.tex
\providecommand{\aut}[1]{\textbf{#1}}
\providecommand{\af}[1]{{\small #1}}

\providecommand{\afn}[1]{\textcolor{antique}{$^{#1}$}}

\togglefalse{goodfireonly}  % flip to \toggletrue{goodfireonly} for Goodfire-only version
\newcommand{\paperauthors}{%
\vspace{-10mm}
\\
\authorentry{Mozes Jacobs}{\costar,\bullet,\extaff{a}} \authorsep
\authorentry{Aiden Swann}{\costar,\bullet,\extaff{b}} \authorsep
\authorentry{Mathilde Papillon}{\costar,\bullet,\extaff{c}} \authorsep 
\\
\authorentry{Siddharth Boppana}{\bullet} \authorsep 
\authorentry{Fenil R. Doshi}{\bullet,\extaff{a}} \authorsep 
\authorentry{Usha Bhalla}{\bullet} \authorsep 
\authorentry{Thomas Icard}{\bullet,\extaff{b}} \authorsep 
\authorentry{Matt Feiszli}{\bullet} \authorsep 
\\
\authorentry{Thomas McGrath}{\bullet} \authorsep
\authorentry{Vasudev Shyam}{\bullet} \authorsep
\authorentry{Demba Ba}{\extaff{a}} \authorsep
\authorentry{Leon Bergen}{\bullet} \authorsep
\authorentry{Nina Miolane}{\extaff{c}} \authorsep
\\
\authorentry{Owen Lewis}{\bullet} \authorsep
\authorentry{Jack Merullo}{\bullet} \authorsep
\authorentry{Ekdeep Singh Lubana}{\bullet} \authorsep
\authorentry{Atticus Geiger}{\bullet} 
\\
\authorentry{Matthew Kowal}{\colead,\bullet} \authorsep
\authorentry{Thomas Fel}{\colead,\bullet} \authorsep \\
\vspace{-2mm}
\\
$^\costar$\af{Equal contribution} \authorsep
$^\colead$\af{Equal senior contribution} \\
\\
\vspace{-6mm}
\\
\goodfireaff 
\\
\repolink{https://github.com/goodfire-ai/vla-geometry}
\\
\vspace{2mm}
\extline{%
  \afflabel{\bullet}{Goodfire}\authorsep
  \afflabel{a}{Harvard University}\authorsep
  \afflabel{b}{Stanford University}\authorsep
  \afflabel{c}{University of California, Santa Barbara}\authorsep
}
\vspace{-2mm}
}

%% file: realization.tex
\newcommand{\realmap}{\bm{\varphi}}

Before specifying the seven concepts of interest that this study will examine and documenting their realized geometry, we first clarify what is meant by realized and by geometry.
In particular, our aim here is to distinguish two questions that are frequently conflated in the literature: whether a concept can be recovered from a policy's activations -- which we term \textit{realized} -- and what geometric shape that recovered concept takes within activation space -- which we term \textit{realized geometry}. 
The first question asks about recoverability, the second about structure; we now formalize those notions.

%%%%%%%%%
\paragraph{Notation.}
Throughout, bold symbols denote vector or matrix valued quantities and calligraphic symbols their domains. 
We model observations as arising from a data-generating process whose latent state collects \(m\) concepts, writing \([m] = \{1,\ldots,m\}\) and \(\bm{z} = (\bm{z}_1,\ldots,\bm{z}_m) \sim p_{\bm{z}}\) for a state drawn from \(\bm{\mathcal{Z}} = \mathcal{Z}_1 \times \cdots \times \mathcal{Z}_m\). 
Borrowing the language of conceptual spaces~\citep{gardenfors2004conceptual}, we take each concept \(i \in [m]\) to be a structured domain \((\mathcal{Z}_i, d_i)\) rather than a single vector, so that \(\mathcal{Z}_i\) supplies the values the concept can assume; a domain may be discrete, continuous, or manifold-valued, with an intrinsic dimension \(k_i = \dim (\mathcal{Z}_i)\) when smooth. 
We write \(\bm{x}\) and \(\bm{a}\) for a single input and activation, and \(\bm{X}\) and \(\bm{A} \in \mathbb{R}^{n \times d}\) for \(n\) of them stacked row-wise.

\paragraph{Data-generating process.}
We consider the minimal data-generating process
\begin{equation}
    \bm{z} \sim p_{\bm{z}},
    \qquad
    \bm{x}
    =
    \bm{g}\left(\bm{z}\right)
    \in \mathcal{X},
    \qquad
    \bm{a}
    =
    \bm{f}\left(\bm{x}\right)
    \in
    \mathcal{A}
    \subseteq
    \mathbb{R}^{d},
    \label{eq:dgp}
\end{equation}
in which the (potentially stochastic) generator \(\bm{g} : \bm{\mathcal{Z}} \to \mathcal{X}\) combines the concepts $\bm{z}$ and instantiates an observation $\bm{x}$ (e.g. an image), while \(\bm{f} : \mathcal{X} \to \mathcal{A}\) maps that observation to an activation $\bm{a}$ at a fixed layer of a VLA. 
In this work, we will focus on a particular set of concepts which are control-relevant and are modeled by both a \textit{quantity} and its \textit{possible values}\footnote{This mirrors the recent notion of a feature field~\citep{yocum2025neural}, where a feature is defined over a topological space -- rather than at a point -- and the geometry of that space embedded in activation space determines which functions of the concept are linearly representable.}, so that \(\mathcal{Z}_i\) may describe a binary gripper state, a scalar distance in \(\mathbb{R}_{+}\), an end-effector position in \(\mathbb{R}^{3}\), or a wrist orientation in \(\mathbb{S}^{1}\).
These are properties of the world or of the control process, and the policy accesses them only through the observation that the generator \(\bm{g}\) produces, so that nothing guarantees a concept present in \(\bm{z}\) remains recoverable from \(\bm{a}\). 
Whether it does is the first thing to establish, and it is a question that shall become well posed only once the means of recovery is fixed.

\begin{definition}[Concept realization]
\label{def:realization}
Let \((\mathcal{Z}_i, d_i)\) be the concept domain of interest and \(\mathcal{V}\) be a class of readouts \(\bm{\gamma} : \mathcal{A} \to \mathcal{Z}_i\), for \(\varepsilon \geq 0\), we say that concept \(i\) is \((\mathcal{V}, \varepsilon)\)-realized in \(\bm{a}\) if
\begin{equation}
    \inf_{\bm{\gamma} \in \mathcal{V}}
    \ \mathbb{E}
    \left(
        d_i \left(
            \bm{\gamma} \left( \bm{a} \right),
            \bm{z}_i
        \right)
    \right)
    \leq
    \varepsilon ,
    \label{eq:realization}
\end{equation}
where the expectation is taken over the process in Equation~\ref{eq:dgp}.
\end{definition}

This definition is relative to both a readout class and an error tolerance. It therefore specifies not only whether information about \(\bm{z}_i\) is present, but also how ``easily'' it can be extracted. 
This follows the same operational principle as predictive \(\mathcal{V}\)-information~\citep{xu2020theory}: usable information is defined relative to a class of predictors.
In particular, unless specified, our main case is the class of affine readouts $\bm{\gamma}\left(\bm{a}\right) = \bm{W}\bm{a} + \bm{b}$.
When concept \(i\) is realized under this class, we say that it is \emph{linearly readable}. 
We reserve \emph{realization} for the general, readout-relative notion and use \emph{linear readability} only for this particular case.

An important remark before jumping to the realized geometry is that realization does not imply that the activations are isomorphic to the concept domain\footnote{It does not require (1) preservation of distances or (2) topology, nor (3) an efficient or minimal representation (see App.~\ref{app:relization_vs_geometry} for an example).} or that the policy causally ``uses'' the recoverable information. Finally, it also lacks a description of the geometry of the conceptual space itself, which we now turn our attention to. 

%\iffalse
\paragraph{Realized geometry.}
Knowing that a concept can be read from the activations tells us nothing about how its values are laid out, and to describe this layout we fix a context \(\bm{c} = \bm{z}_{[m]\setminus\{i\}}\) containing all the other concepts and look at how the activations move when only \(\bm{z}_i\) changes.

\begin{definition}[Realized geometry]
\label{def:realized_geometry}
The activation map of concept \(i\) in context \(\bm{c}\) is
\begin{equation}
    \realmap_i(\bm{v};\bm{c})
    =
    \mathbb{E}\left(
        \bm{a} \mid \bm{z}_i = \bm{v},\, \bm{z}_{[m]\setminus\{i\}} = \bm{c}
    \right),
    \qquad
    \bm{v} \in \mathcal{Z}_i,
    \label{eq:realmap}
\end{equation}
and the realized geometry of the concept is the image \(\mathcal{M}_i(\bm{c}) = \realmap_i(\mathcal{Z}_i;\bm{c})\) together with the map that produces it.
\end{definition}

In words, the realized geometry is the path traced by the activations when only concept \(i\) moves, for example two points for the gripper state, a curve for a distance, or an arc for the wrist angle.
This path is what just monitoring would overlook, since, for example a linear probe only looks at a few directions of activation space and the path can take any shape in the others without changing its accuracy (App.~\ref{app:relization_vs_geometry}).
The rest of this work estimates \(\realmap_i\), directly through controlled sweeps when the simulator lets us move one concept and hold the rest fixed, and through teleports and natural rollouts when it does not, and then tests with interventions whether the policy actually uses the geometry we find.

%% file: extended_results.tex
\section{Extended Results}
\label{app:extended_results}

This appendix collects the figures that support the main text; Figure~\ref{fig:variables} gives an overview of the seven task variables, and experimental details are in Appendix~\ref{app:details}.

\begin{figure}[h!]
    \centering
    \includegraphics[width=0.9\linewidth]{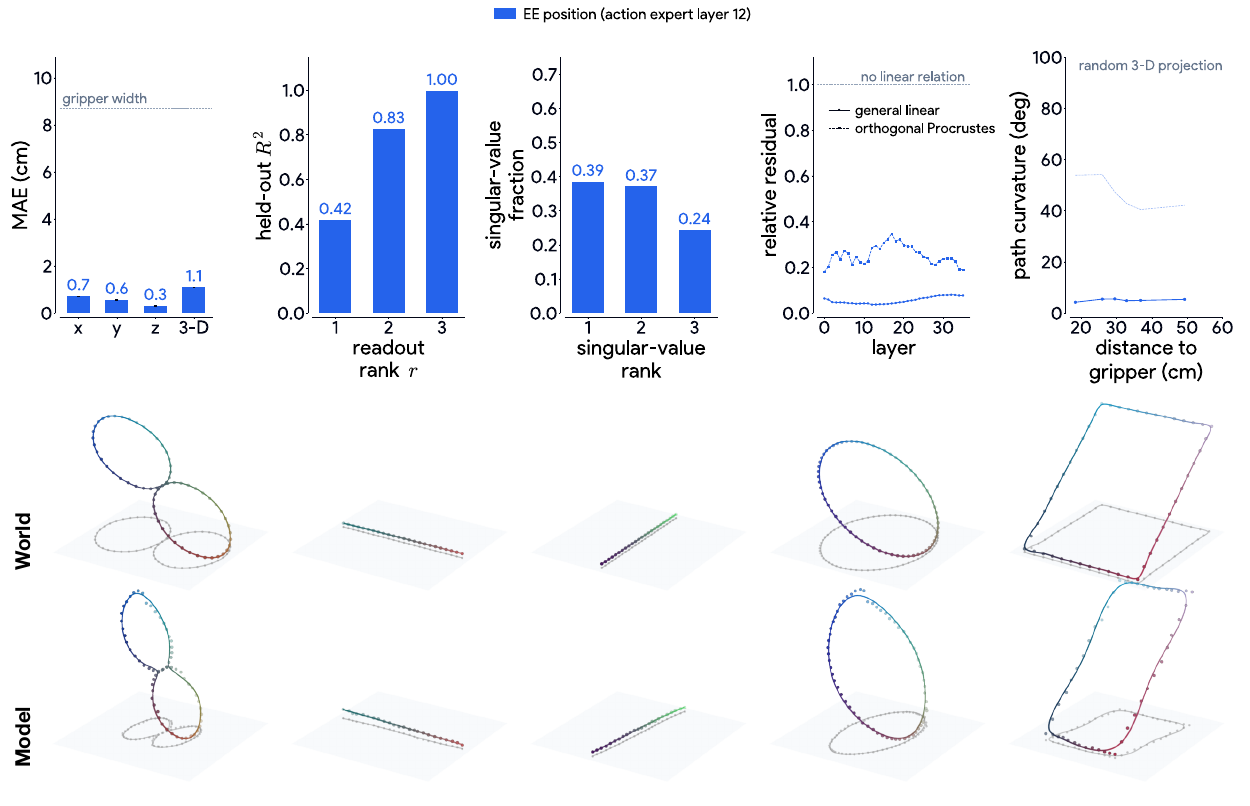}
    \caption{\textbf{The action expert (layer 12) also maintains a rank-3 EE position subspace.}
    Panels as in the top and middle rows of Figure~\ref{fig:position}, for EE position read from the action expert, which receives proprioception; the paths are from a single scene excluded when fitting the subspace. The readout is near-exact and the paths are recovered almost without distortion. The gap between the general-linear and orthogonal residuals shows the subspace is a scaled, not isometric, copy of world coordinates.
    }
    \label{fig:position_ae}
\end{figure}

\begin{figure}[h]
    \centering
    \includegraphics[width=0.9\linewidth]{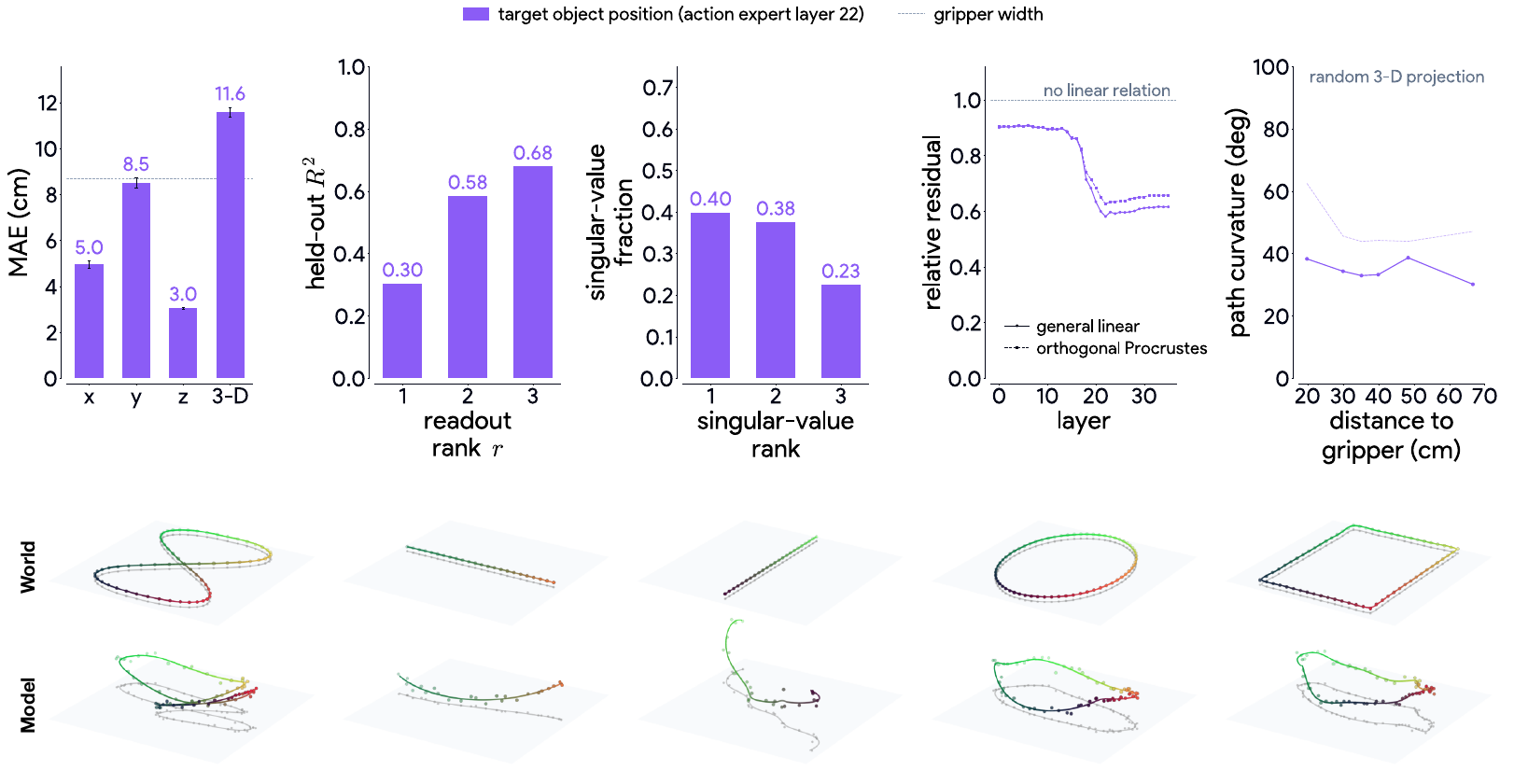}
    \caption{\textbf{The action expert (layer 22) also maintains a rank-3 target object position subspace.}
    Panels as in Figure~\ref{fig:position_ae}, for the target object's position; the paths are from a single scene excluded when fitting the subspace. The action expert receives the object only through cross-attention to the VLM, and it carries target position with larger error than the VLM (Figure~\ref{fig:position}), recovering the paths with more distortion.
    }
    \label{fig:target_pos_ae}
\end{figure}

\begin{figure}[tb]
    \centering
    \includegraphics[width=0.9\linewidth]{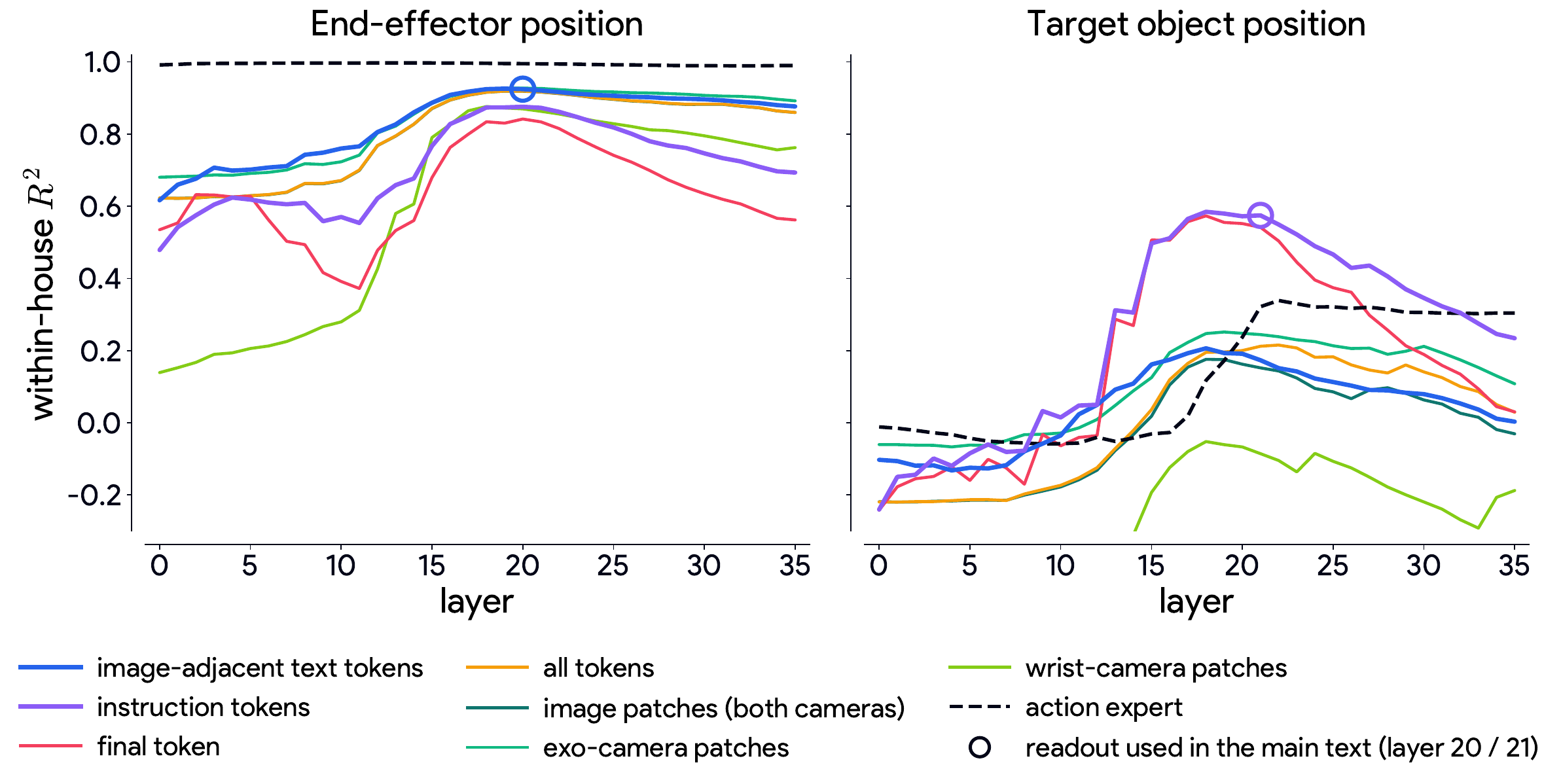}
    \caption{\textbf{Within a scene, target object position is computed on the VLM's instruction and final tokens between layers 12 and 21 and appears in the action expert from layer 18; EE position is present from the first layer in both.}
    Held-out $R^2$ of the rank-3 ridge readout of end-effector position (left) and target object position (right) at every layer, scored after removing each held-out house's mean from the true and predicted positions, so only position changes within a scene count; one line per VLM token pool, the action expert dashed, rings on the readouts of Figure~\ref{fig:position}, values below the axis floor clipped.
    The target's height is constant within a scene, so the target panel's variance is horizontal and any within-scene variation in the predicted height counts as error; scores fall below zero where a readout's predictions vary within a scene without tracking the object.
    On the instruction and final tokens the target read stays at or below zero until layer 12 and peaks at layers 18--21, while the image patches carry little of it at any layer.
    The action expert tracks EE position near-perfectly at every layer and the target only from layer 18 on.}
    \label{fig:position_layer_profiles}
\end{figure}

\begin{figure}[tb]
    \centering
    \includegraphics[width=\linewidth]{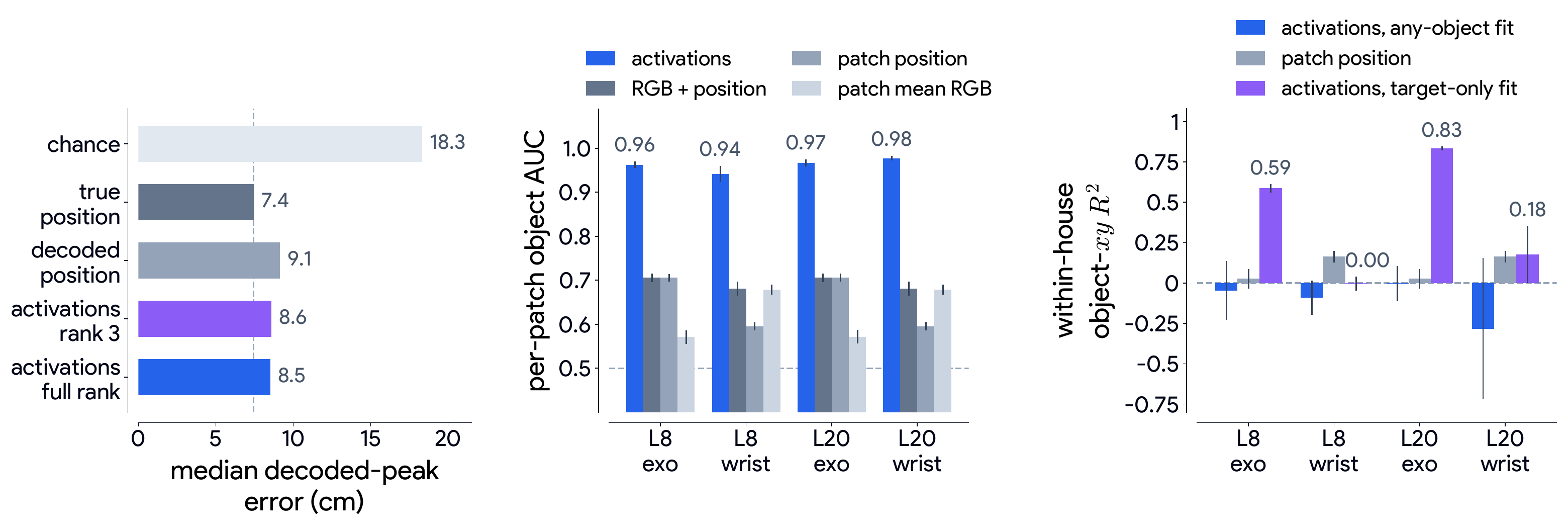}
    \caption{\textbf{The VLM keeps the target's position but no map of the workspace; other objects are encoded per image patch.}
    (Left) A table-occupancy map of the target, read from VLM activations (last token, layer 21), is no more accurate than the same readout applied to the target's three true coordinates (dashed). (Middle) One readout shared across all image-patch tokens detects whether an object the instruction does not name is visible in that patch, well above patch-position and patch-color baselines (exo and wrist cameras, layers 8 and 20). (Right) With scene means removed, a readout of table position fit on all visible objects does not track the target as it moves, whereas one fit on the instructed target alone does on the exo camera.}
    \label{fig:workspace_map}
\end{figure}

\paragraph{End-effector and target object position (App.~\ref{app:position}).}
Figure~\ref{fig:crossed_store_map} shows the crossed arm${\times}$object teleports used for both position variables: unlike the object teleports, both the arm and the object move, so the horizontal positions of the two vary nearly independently on the same states.
The action expert also carries rank-3 subspaces for EE position (Figure~\ref{fig:position_ae}) and target position (Figure~\ref{fig:target_pos_ae}); the EE subspace is a stretched rather than rigid copy of world coordinates, and the target, which reaches the action expert only through cross-attention to the VLM, is decoded with larger error than in the VLM.
Figure~\ref{fig:position_layer_profiles} marks the token groups and layers used in Figure~\ref{fig:position}: within a scene, EE position is decodable from the first layer, whereas target position appears on the instruction and final tokens between layers 12 and 21, and in the action expert only from layer 18.
Figure~\ref{fig:workspace_map} shows that the VLM's target code is a position rather than a map of the workspace: an occupancy map read from the activations is no more accurate than one built from the target's true coordinates, objects the instruction does not name are detected per image patch, and a table-position readout fit on all visible objects does not track the target as it moves within a scene, whereas one fit on the target alone does on the exo camera.

\begin{figure}[tb]
    \centering
    \includegraphics[width=0.75\linewidth]{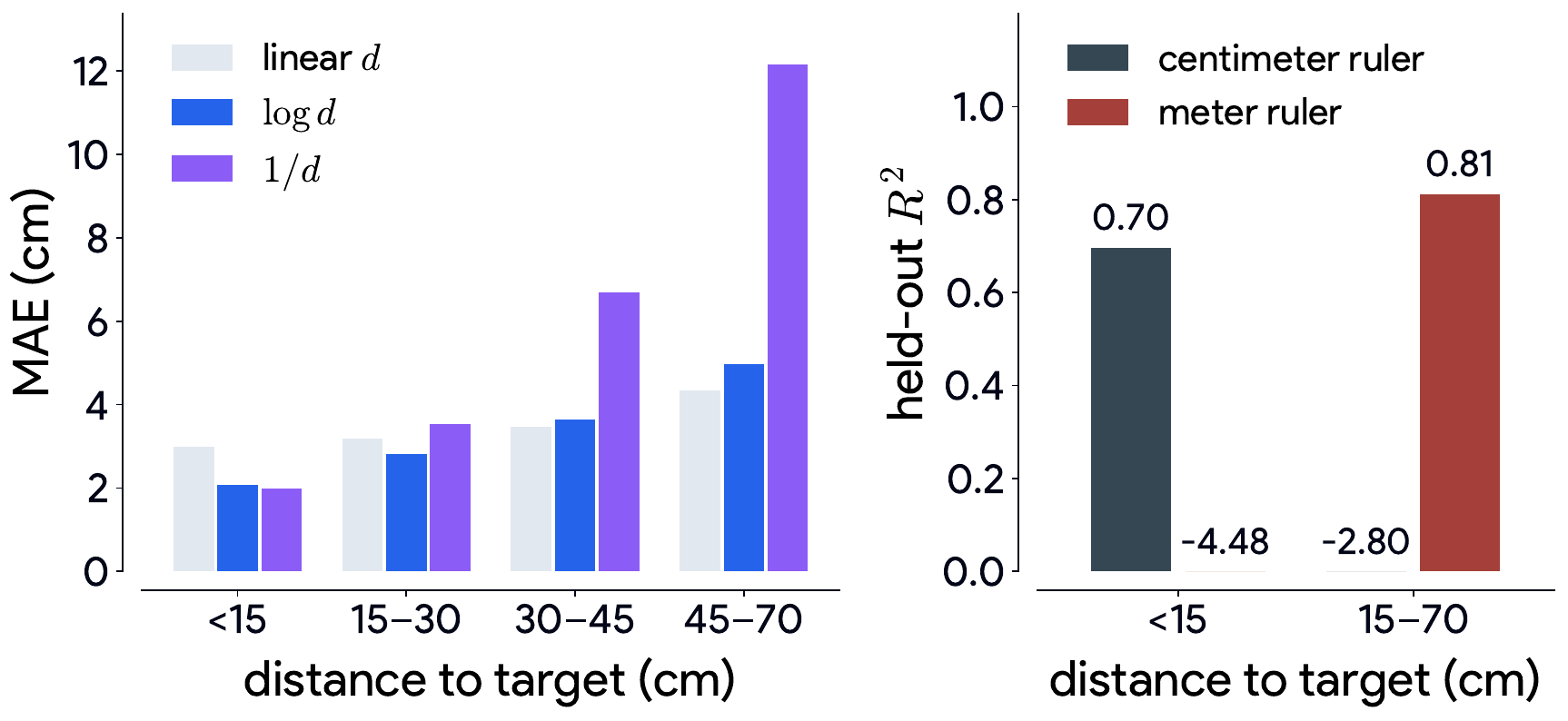}
    \caption{\textbf{Distance is compressive within the near range and closest to linear beyond 30\,cm, and each ruler reads only its own range (VLM, layer 21).}
    (Left) Held-out mean absolute error per bin of true distance for a single full-range ridge probe fit to $d$, $\log d$, or $1/d$ and mapped back to metres; the compressive links have the lower error below 15\,cm and the linear link in the farthest bins.
    (Right) Held-out $R^2$ against true distance for the centimeter ruler and the meter ruler of Figure~\ref{fig:dist_target}, scored separately on states below and above that cutoff; negative (saturated) values are drawn at zero and printed.
    Each ruler saturates outside its own range, as does a single unclipped metre-scale probe (not shown) on the near-range states.
    }
    \label{fig:distance_link_race}
\end{figure}

\begin{figure}[tb]
    \centering
    \includegraphics[width=0.75\linewidth]{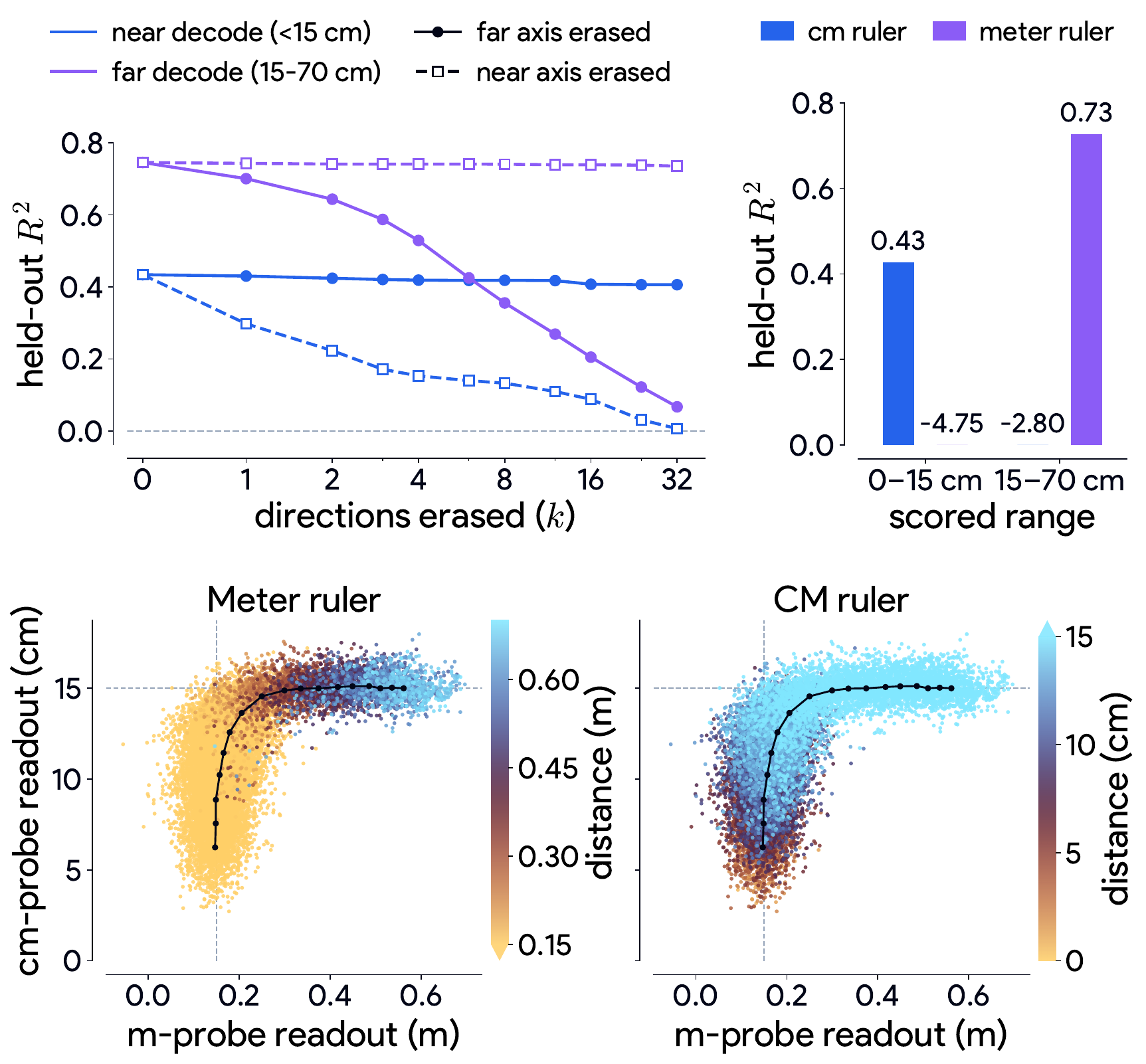}
    \caption{\textbf{The action expert (layer 35) reproduces the two-subspace distance code.}
    Same store, folds, and constructions as Figure~\ref{fig:dist_target}, read from the action expert instead of the VLM.
    (Top left) Held-out $R^2$ of the near and far decodes as far-axis (solid) or near-axis (dashed) directions are erased; each decode drops only when its own axis is removed.
    (Top right) Held-out $R^2$ of the centimeter and meter rulers on each side of the 15\,cm cutoff; each ruler scores only its own range, and bars below zero are drawn at zero with the value printed.
    (Bottom) The two ruler readouts plotted against each other, colored by true distance on the meter (left) and centimeter (right) scales; the meter ruler resolves the far arm and the centimeter ruler the near arm, as in the VLM cell.
    }
    \label{fig:distance_ae35}
\end{figure}

\begin{figure}[t]
    \centering
    \includegraphics[width=0.5\linewidth]{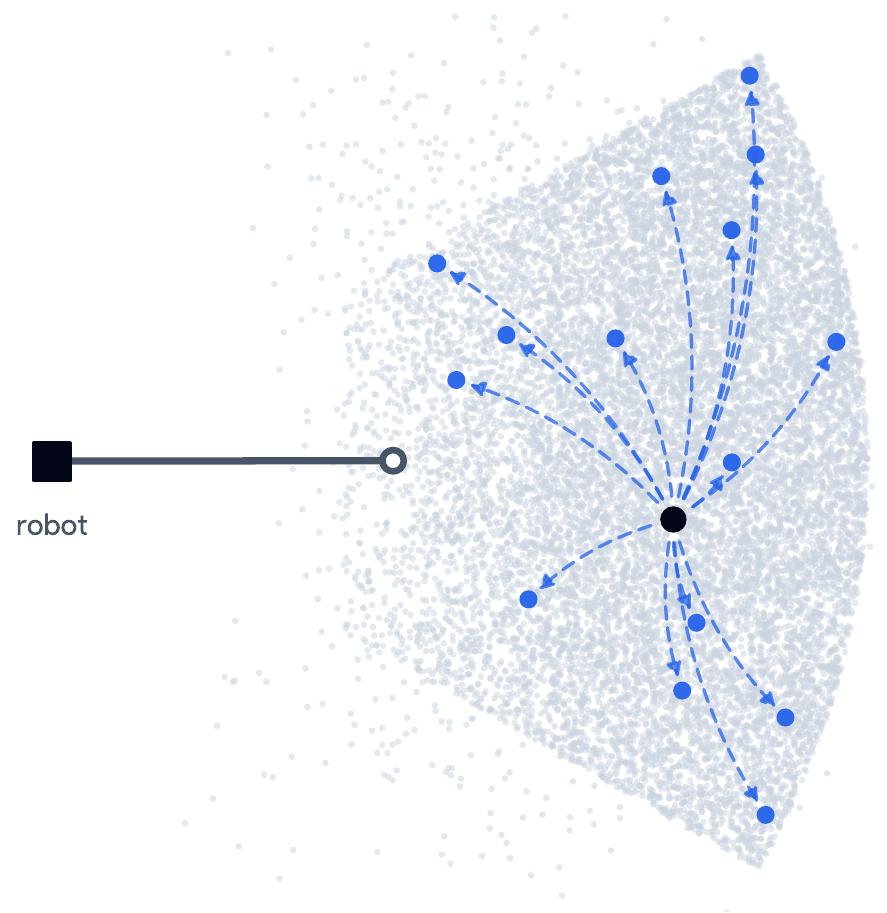}
    \caption{\textbf{Object teleportation across the workspace for distance to target object experiments.} Object placements in the robot-base frame, viewed from above (robot at left, forward to the right). Gray points are all placements in the teleport datasets (906 houses). For one example house, the black point is the object's natural placement and the blue points its 15 teleported placements, sampled uniformly over the supporting surface within the benchmark placement envelope; dashed arrows connect the natural placement to each teleport. Only the object is moved; the arm's initial pose is unchanged.}
    \label{fig:teleport_map}
\end{figure}

\begin{figure}[t]
    \centering
    \includegraphics[width=\linewidth]{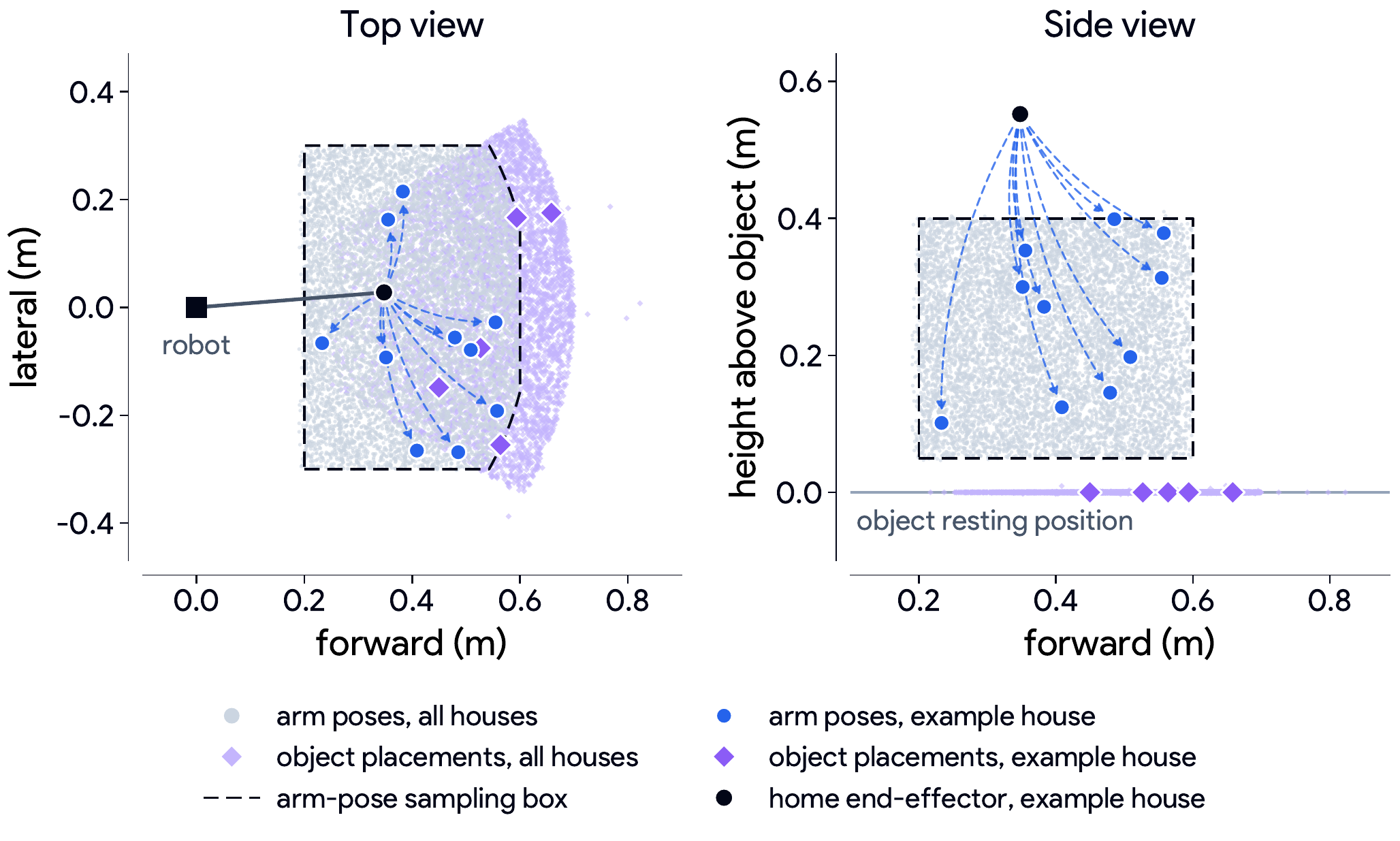}
    \caption{\textbf{Arm and object placements in the crossed teleport harvest used for EE and target position.}
    Robot-base frame viewed from above (left) and from the side, with height above the object's resting position (right). Gray points are arm poses and violet diamonds object placements across all houses; the dashed outline is the box arm poses are sampled in. For one example house, the black point is the home pose, blue points its ten arm poses, and violet diamonds its five object placements; unlike Figure~\ref{fig:teleport_map}, both the arm and the object move.}
    \label{fig:crossed_store_map}
\end{figure}

\paragraph{Distance to target object (App.~\ref{app:distance}).}
Figure~\ref{fig:teleport_map} shows the object placements in the object-teleport dataset used for distance and direction; only the object is teleported.
Figure~\ref{fig:distance_link_race} shows that a single full-range probe is most accurate with a compressive link ($\log d$ or $1/d$) below 15\,cm and with a linear link in the farthest bins, and that the centimeter and meter rulers each score only on their own side of the cutoff.
Figure~\ref{fig:distance_ae35} repeats the erasure test and both rulers at the action expert's last layer and finds the VLM's two-subspace code.

\begin{figure}[tb]
    \centering
    \includegraphics[width=\linewidth]{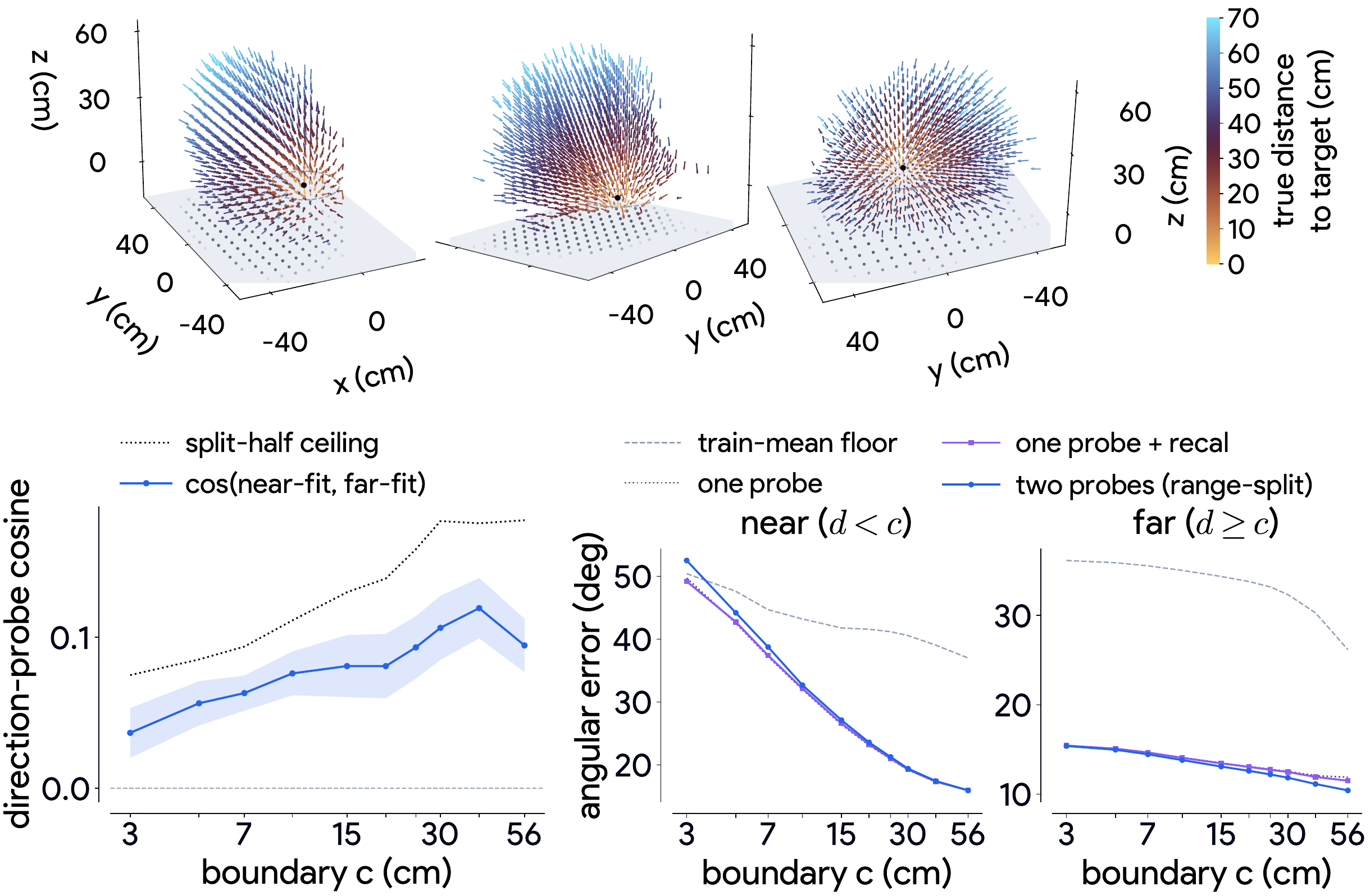}
    \caption{\textbf{Direction to target object is observable with one probe, in contrast to the two distance rulers.}
    \textbf{Top:} held-out decoded directions (VLM layer 21), averaged in $6$\,cm cells in a target-centered frame; arrow length is cell coherence. Decoded directions point toward the target throughout the workspace (median angular error $11.7^{\circ}$ vs.\ a $33.1^{\circ}$ floor); within $15$\,cm, per-read error increases but cell means still point toward the target (coherence $0.98$ beyond $40$\,cm vs.\ $0.77$ inside $7.5$\,cm).
    \textbf{Bottom left:} as in Fig.~\ref{fig:dist_target}c, but for direction probes fit on different halves of the houses: the near/far cosine stays below its ceiling at every boundary ($0.081$ vs.\ $0.130$ at $c=15$).
    \textbf{Bottom right:} the range-split pair does not outperform a single recalibrated probe on near reads (up to $3^{\circ}$ worse) and improves on it by at most $1.1^{\circ}$ on far reads --- a single probe suffices at all ranges.}
    \label{fig:direction_vlm}
\end{figure}

\paragraph{Direction to target object (App.~\ref{app:direction}).}
Figure~\ref{fig:direction_vlm} shows that, unlike distance, direction to the target is read by a single probe: decoded directions point toward the target across the workspace, and splitting the probe by range does not improve near reads and improves far reads by at most $1.1^\circ$.

\begin{figure}[tb]
    \centering
    \includegraphics[width=\linewidth]{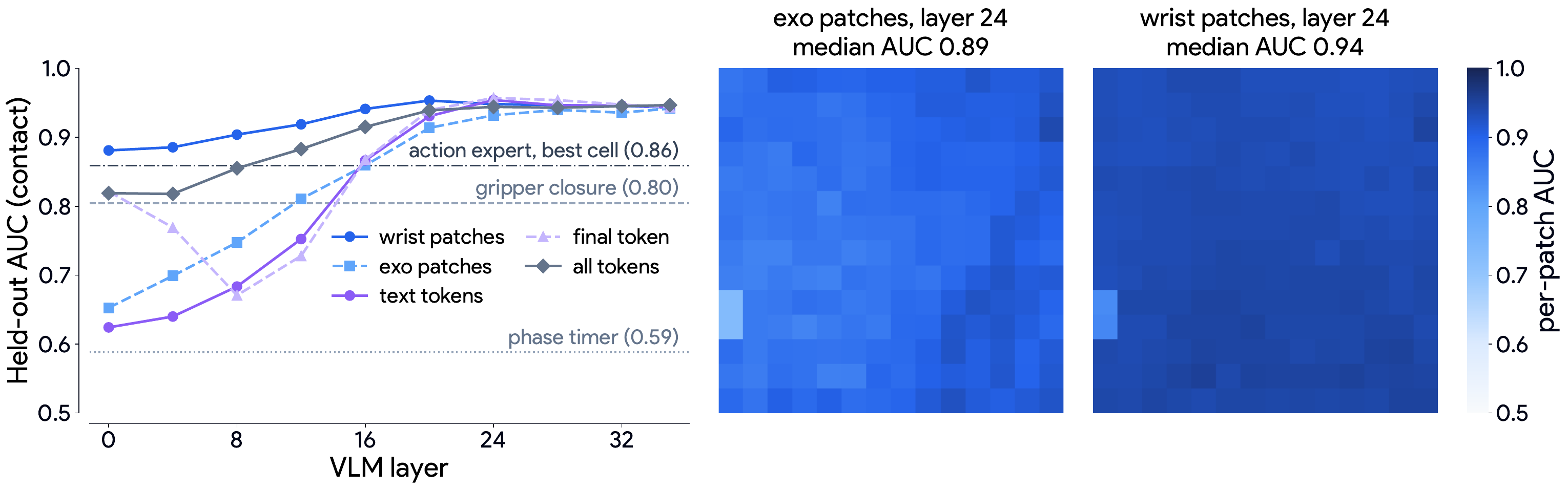}
    \caption{\textbf{Contact is decodable from the wrist-camera patches at the first layer and from the exo-camera patches and text tokens at a similar level only from the middle layers on (VLM).}
    (Left) Held-out AUC of the contact discriminant at each of 10 layers, fit separately on five token groups: wrist-camera patches, exo-camera patches, text tokens, the final token, and the whole-sequence mean; reference lines mark a phase-timer baseline, a gripper-closure baseline, and the action expert.
    (Right) Per-patch AUC at layer 24 for the two cameras: the signal is spread over the whole image rather than confined to the fingers.}
    \label{fig:contact_token_layer_map}
\end{figure}

\paragraph{Contact (App.~\ref{app:contact}).}
Figure~\ref{fig:contact_token_layer_map} shows where in the VLM contact is decodable: from the wrist-camera patches already at the first layer, while the exo-camera patches and text tokens reach a similar level only in the middle layers; at layer 24, the per-patch signal covers the whole image rather than only the fingers.

\begin{figure}[tb]
    \centering
    \includegraphics[width=\linewidth]{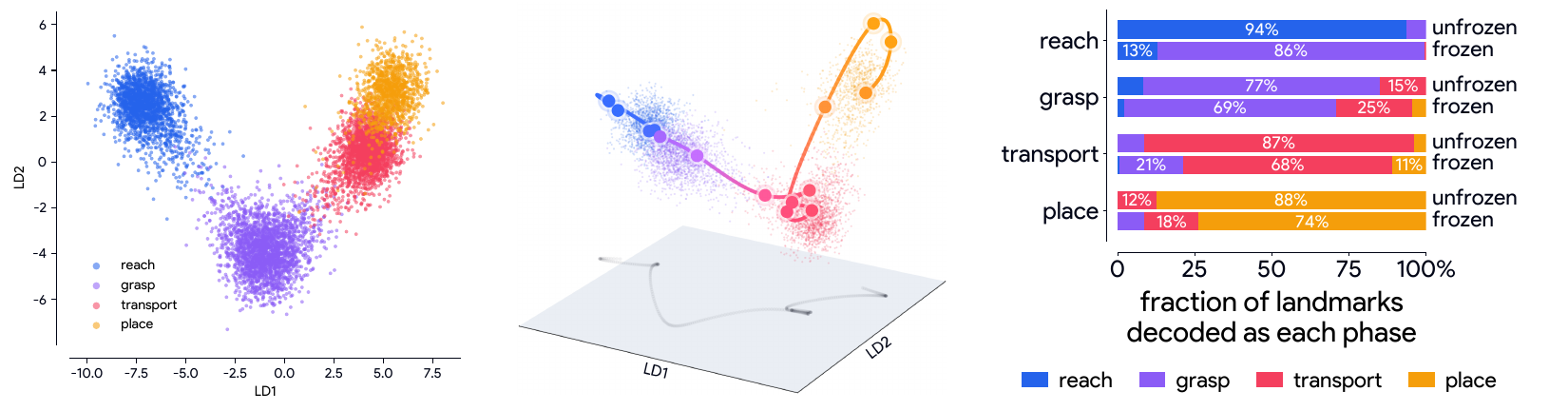}
    \caption{\textbf{MolmoBot's task phase subspace in the VLM backbone (vision tokens, Layer 31).}
    (Left) LDA projection of natural rollouts, colored by phase (0.86 balanced accuracy across held-out houses; AE: 0.92).
    (Middle) The full three-dimensional LDA subspace, with a held-out episode's trajectory traversing the four phase regions in order. 
    (Right) Freezing shifts the VLM's decoded phase toward grasp, as in the action expert (Figure~\ref{fig:phase}): frozen reach and grasp landmarks are mostly decoded as grasp, and frozen transport and place landmarks are most often still decoded as transport and place.
    }
    \label{fig:phase-vlm}
\end{figure}

\begin{figure}[tb]
    \centering
    \includegraphics[width=\linewidth]{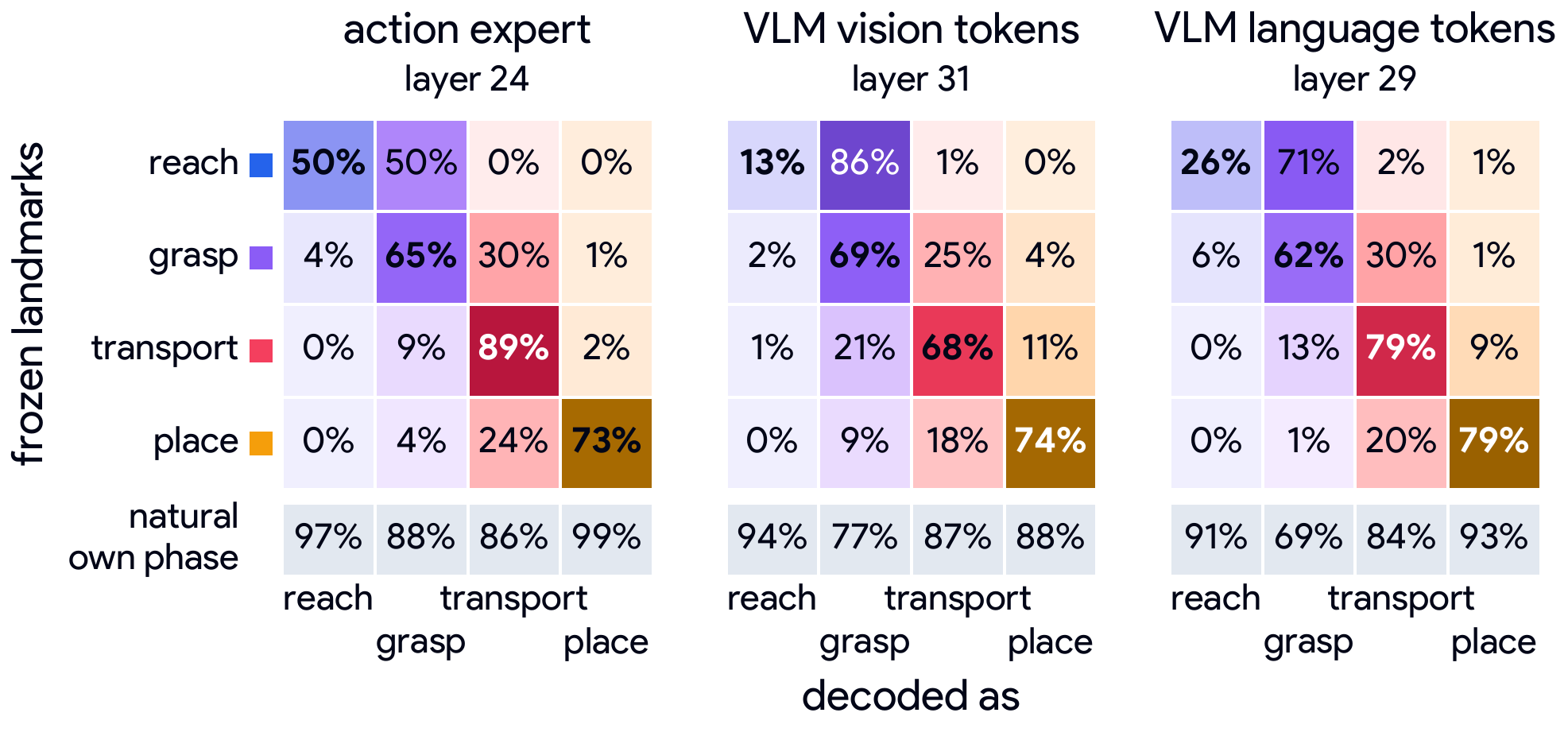}
    \caption{\textbf{Under the frozen read, the decoded phase shifts toward grasp in all three streams, while place landmarks are most often still decoded as place.}
    Left to right: action expert, VLM vision tokens, VLM language tokens. Each row gives the percentage of one phase's landmarks decoded as each phase when the policy is shown the current frame twice (bold: the true phase); the strip beneath each matrix gives the percentage decoded correctly with the usual two frames, at the same moments.}
    \label{fig:phase-frozen-matrices}
\end{figure}

\begin{figure}[h]
    \centering
    \includegraphics[width=0.75\linewidth]{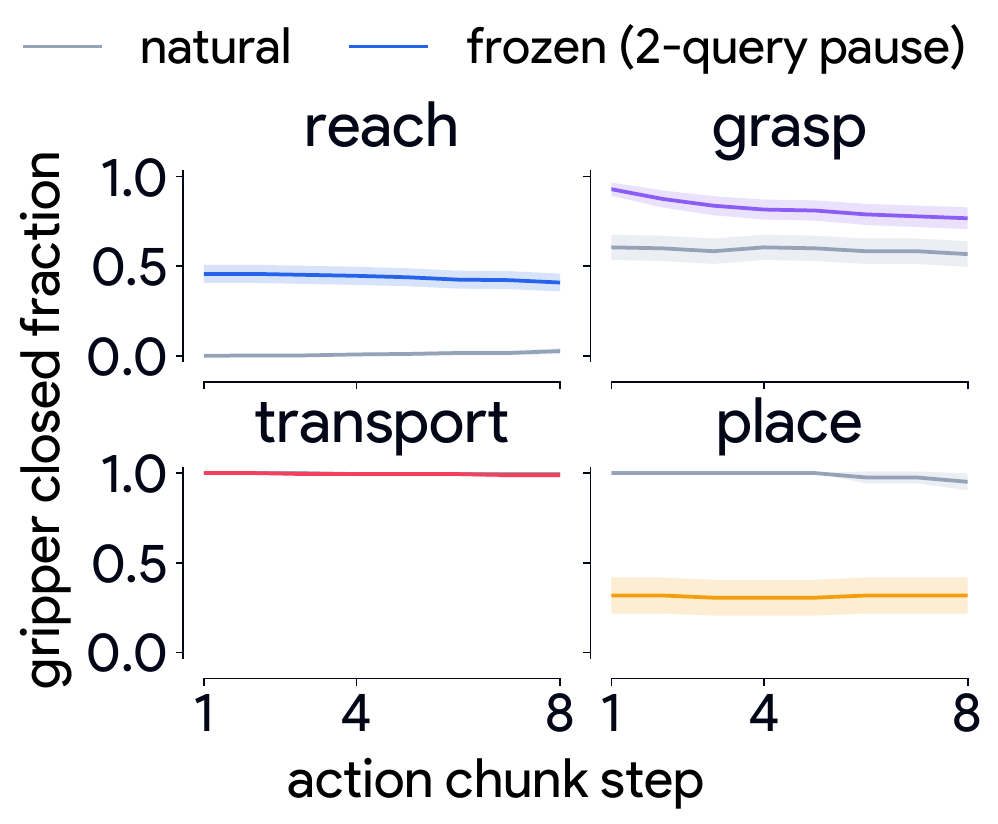}
    \caption{\textbf{Per-step time course of the freezing effect on gripper commands.}
    Fraction of chunks commanding a closed gripper at each step of the executed action chunk (shaded: 95\% CI), natural vs.\ frozen, per phase.
    }
    \label{fig:phase-freeze}
\end{figure}

\paragraph{Phase (App.~\ref{app:phase}).}
Figure~\ref{fig:phase-vlm} shows the phase subspace in the VLM's vision tokens: the four phases occupy separate regions in temporal order, and freezing shifts the VLM's decoded phase toward grasp, as in the action expert.
Figure~\ref{fig:phase-frozen-matrices} compares the frozen read across the action expert and both VLM token groups: in all three, the decoded phase shifts toward grasp, while place landmarks are most often still decoded as place.
Figure~\ref{fig:phase-freeze} gives the per-step time course of the freezing effect on gripper commands in each phase.

\begin{figure}[tb]
    \centering
    \includegraphics[width=\linewidth]{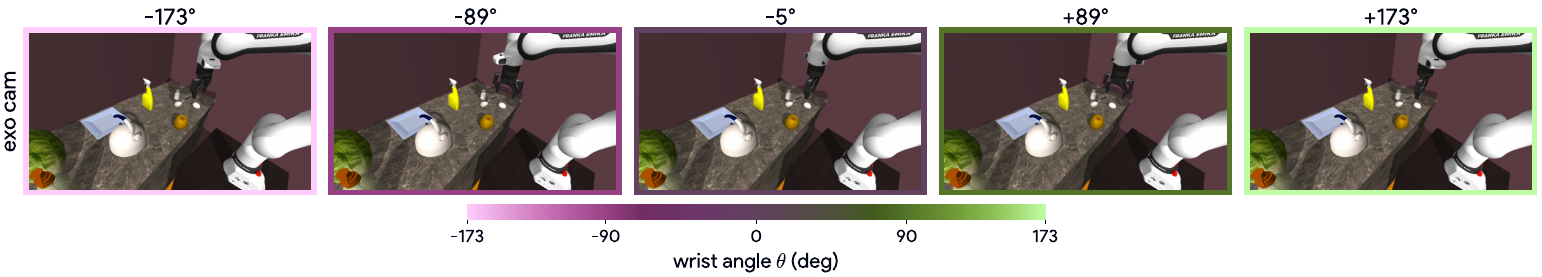}
    \caption{\textbf{Wrist roll sweep (Figure~\ref{fig:wrist}).} Exo (third-person) camera images at several points of the sweep through the wrist's full roll range ($-172.8^\circ$ to $+172.8^\circ$), with all other joints fixed. Borders are colored by wrist angle $\theta$.}
    \label{fig:wrist_frames}
\end{figure}

\begin{figure}[tb]
    \centering
    \includegraphics[width=\linewidth]{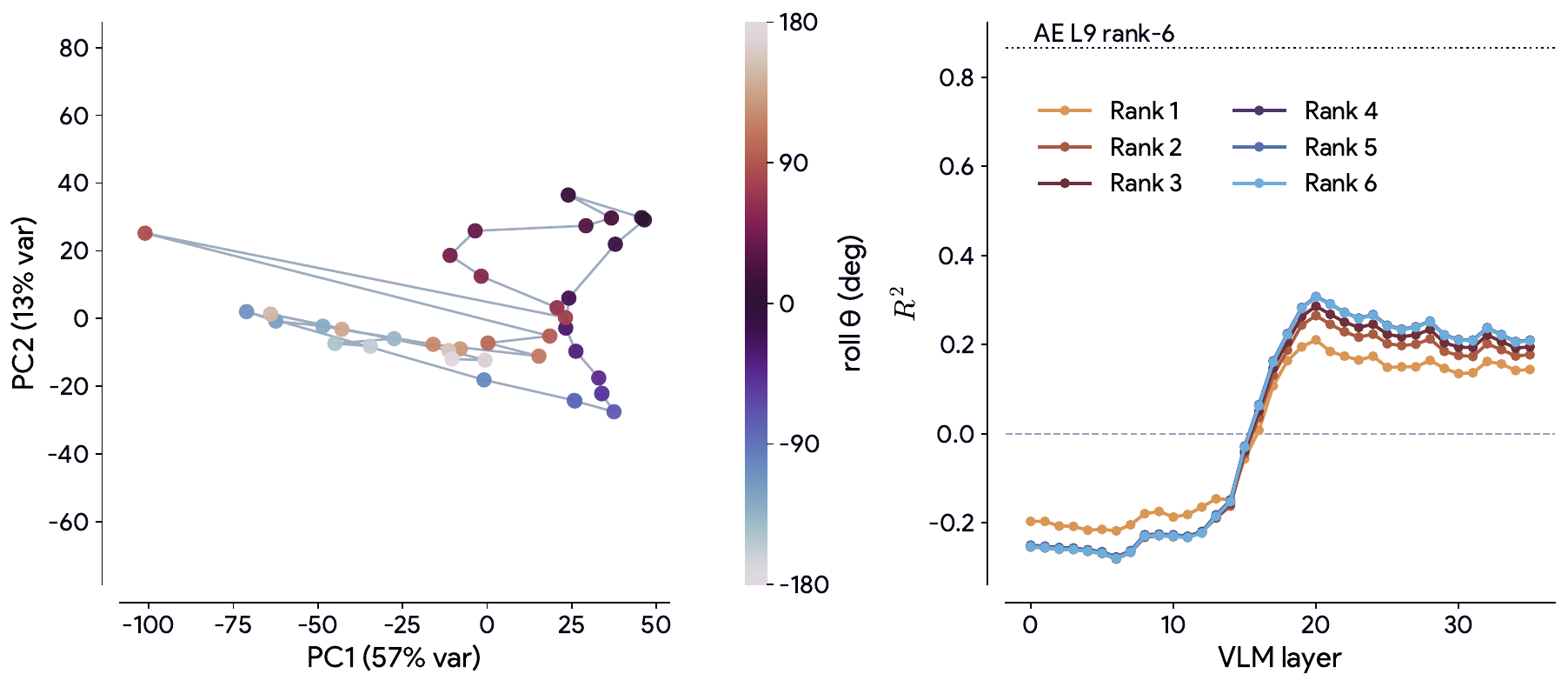}
    \caption{\textbf{The VLM maintains only a weak, pixel-derived trace of wrist orientation.}
    A roll sweep from the same starting pose as Figure~\ref{fig:wrist}, over a $332^\circ$ range ($-156^\circ$ to $+176^\circ$), read at the VLM; the VLM receives only images and text as input.
    (Left) PCA over the sweep activations (layer 35). The trace wraps --- $-156^\circ$ and $+176^\circ$ land close together --- but does not form a clean circle.
    (Right) The reduced-rank regression of Figure~\ref{fig:wrist_rank} at every VLM layer, on natural rollouts re-encoded at the VLM (19{,}089 states, 169 houses). Decoding peaks mid-stack at $R^2 = 0.31$ (layer 20), a third of the action expert's $0.87$ (dotted) on the same data.
    }
    \label{fig:wrist_vlm}
\end{figure}

\begin{figure}[tb]
    \centering
    \includegraphics[width=\linewidth]{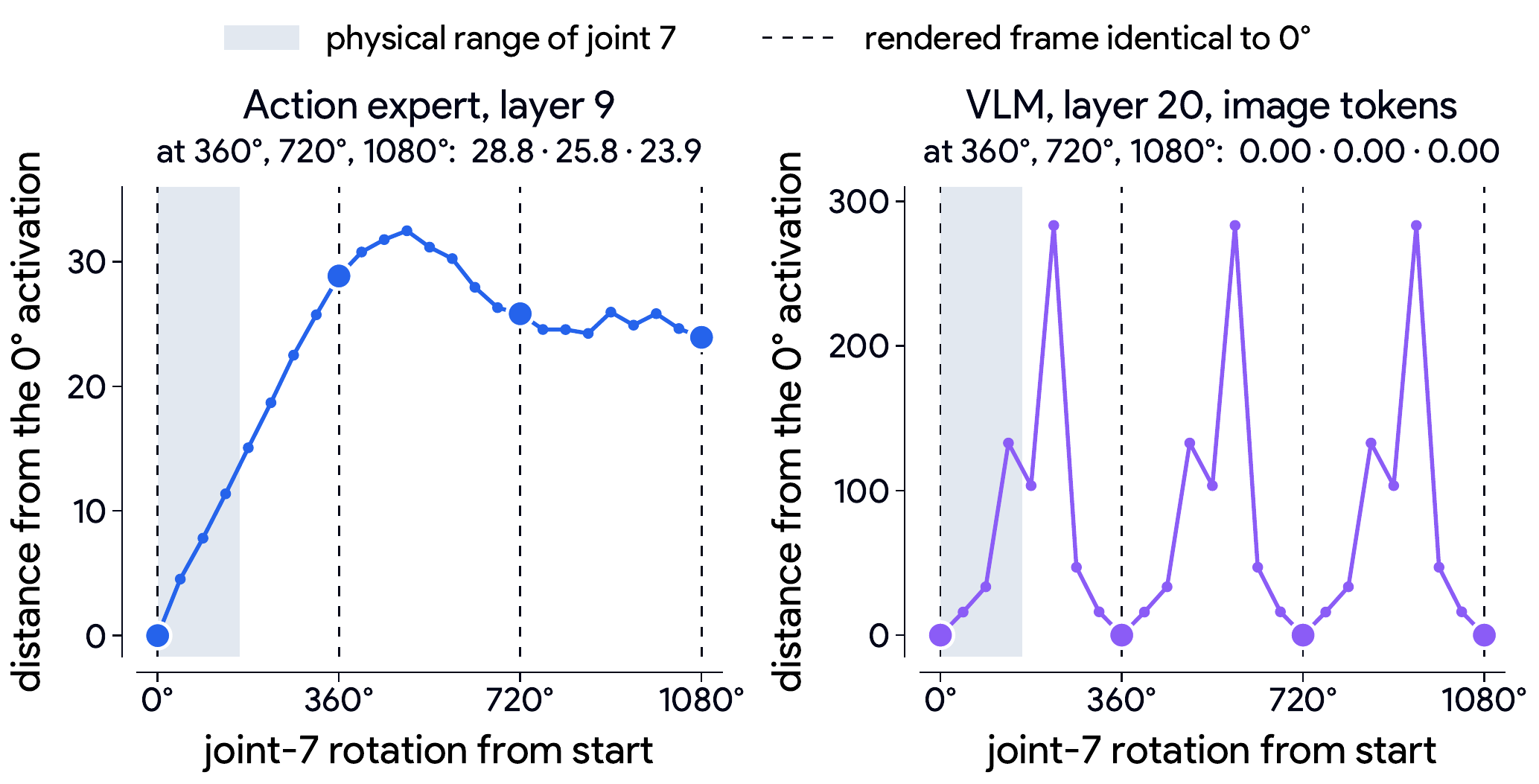}
    \caption{\textbf{After a full turn of the wrist joint the action expert's activation does not return to its start; the VLM's returns exactly.}
    From one pose of the roll sweep (App.~\ref{app:wrist}; a single house), the wrist joint (Franka joint~7) is driven through successive full turns by writing the joint value directly into the simulator, far past the joint's physical range (shaded band), so the sweep tests the models' inputs rather than a reachable motion.
    At every full turn (dashed lines, large markers) the rendered frames and text tokens are identical to the start and only the raw joint value differs, and only the action expert receives that value.
    \textbf{(Left)} Action expert, layer~9, first action token: Euclidean distance of the activation at each step from the start activation; after one turn it is $28.8$ units away and does not return over the following turns.
    \textbf{(Right)} VLM, layer~20, mean over image tokens: the same distance, which varies within each turn and is exactly $0$ at every full turn; the two panels are in different activation units.
    }
    \label{fig:wrist_multiturn}
\end{figure}

\begin{figure}[tb]
    \centering
    \includegraphics[width=\linewidth]{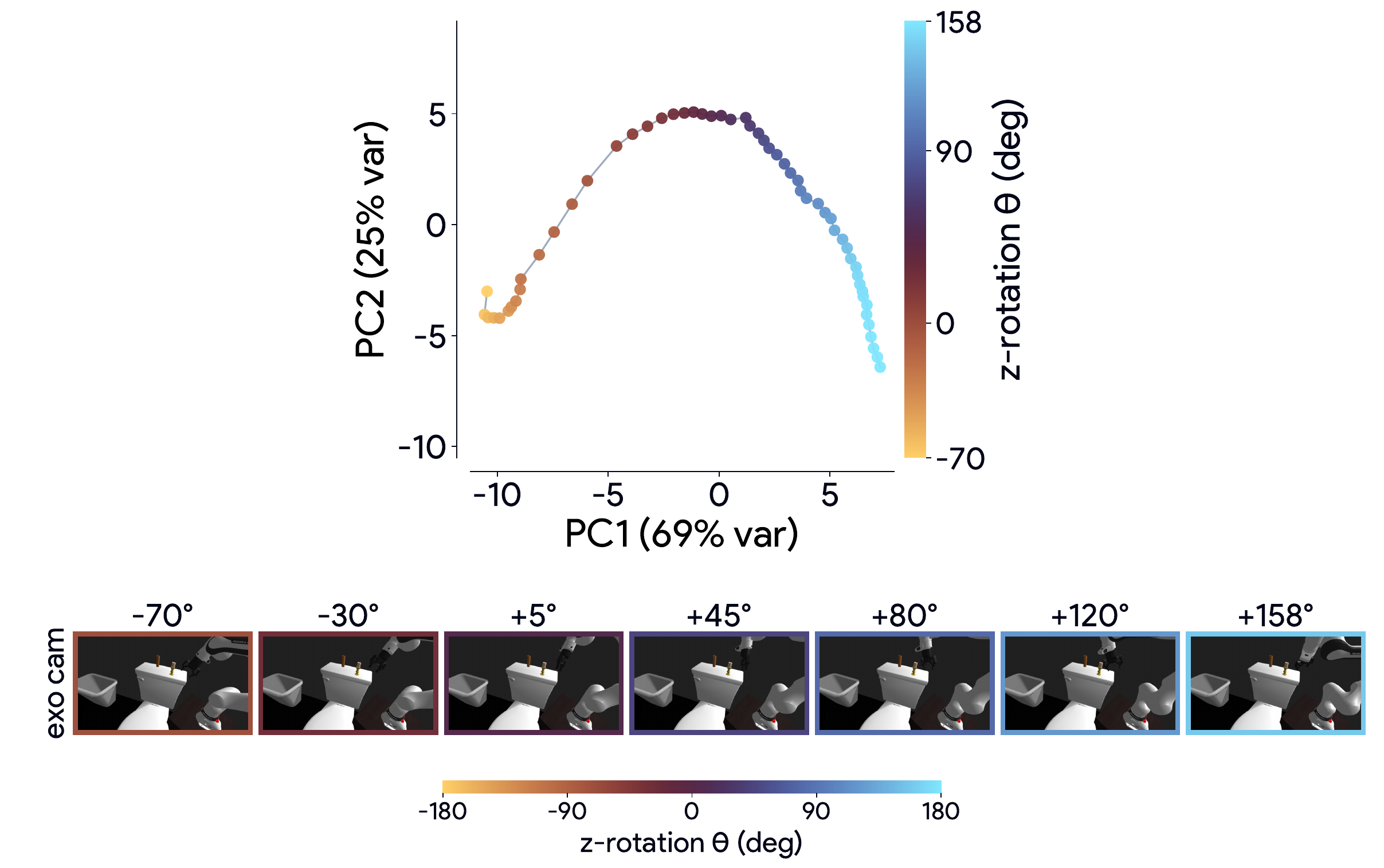}
    \caption{\textbf{Even the most circular reachable wrist rotation does not close.}
    Instead of rolling about the gripper's approach axis, we tilt the gripper
    $30^\circ$ off that axis and orbit the tilt about the vertical. If driven through a full turn, this precession returns the wrist orientation exactly to its start (a closed circle in $SO(3)$). The gripper position is held fixed within 5\,cm while the whole arm reconfigures, and we record activations at each step. The arm's joint limits (clipped to the 1st to 99th training percentile) stop the orbit partway: sweeping both directions from the start covers $-70^\circ$ to $+158^\circ$ ($228^\circ$ of the turn; angles are the realized rotation about the vertical axis).
    (Top) PCA over the sweep activations. Precession remains an open arc, with PC1 carrying $69\%$ of the variance and PC2 $25\%$.
    (Bottom) The exo camera image at different rotation angles along the sweep.
    }
    \label{fig:wrist_circle}
\end{figure}

\begin{figure}[tb]
    \centering
    \includegraphics[width=0.6\linewidth]{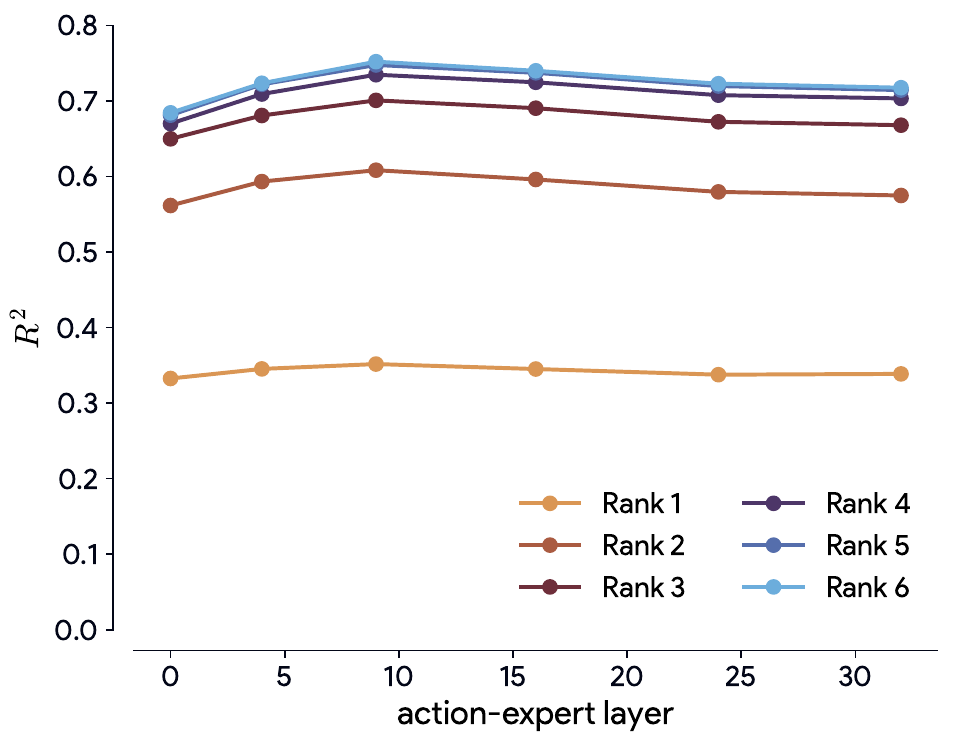}
    \caption{\textbf{Full 3-DOF wrist orientation is decodable from natural rollouts at low rank (AE).}
    Unlike Fig.~\ref{fig:wrist}, which rotates a single roll joint and reads the code with PCA, here ``wrist orientation'' means the complete 3-DOF orientation of the hand, represented by the first two columns of its rotation matrix (a 6-D target).
    We fit reduced-rank ridge regressions of rank 1 to 6 from the activations to this target on natural rollouts (6{,}555 queries, 191 houses) and report $R^2$ on held-out houses at six action-expert layers.
    Rank 2 recovers most of the decodable variance and rank 3 nearly all of it; higher ranks add little. Decodability peaks at layer 9.
    }
    \label{fig:wrist_rank}
\end{figure}

\paragraph{Wrist orientation (App.~\ref{app:wrist}).}
Figure~\ref{fig:wrist_frames} shows the exo-camera view of the roll sweep in Figure~\ref{fig:wrist}.
Figure~\ref{fig:wrist_multiturn} shows that after a full turn of the wrist joint, when the images are identical to the start, the VLM's activation returns to its start but the action expert's does not, consistent with only the action expert receiving the raw, non-wrapping joint value.
Figure~\ref{fig:wrist_vlm} shows that the VLM, which receives only images and text, carries a weaker trace of wrist orientation, with about a third of the action expert's $R^2$.
Figure~\ref{fig:wrist_circle} shows that the action expert's trace stays an open arc over the $228^\circ$ of a precession the arm covers (a full turn would return the orientation to its start).
Figure~\ref{fig:wrist_rank} shows that the full 3-DOF orientation of the hand is decodable from natural rollouts at low rank, best at layer 9.

\begin{figure}[h]
    \centering
    \includegraphics[width=0.5\linewidth]{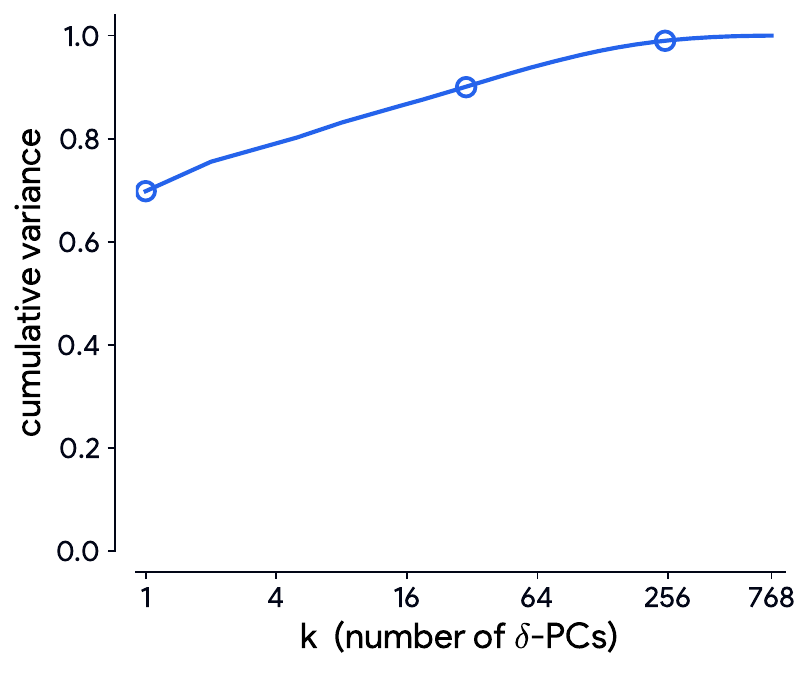}
    \caption{
    \textbf{Cumulative variance of gripper steering vectors in layer 32.}
    Cumulative explained variance from PCA on the pullback steering vectors $\delta$ (App.~\ref{app:gripper}) as a function of the number of principal components. The first component captures ${\sim}70\%$ of the variance, complementing the per-component view in Figure~\ref{fig:gripper}.
    }
    \label{fig:gripper_cumulative_var}
\end{figure}

\begin{figure}[t]
    \centering
    \includegraphics[width=\linewidth]{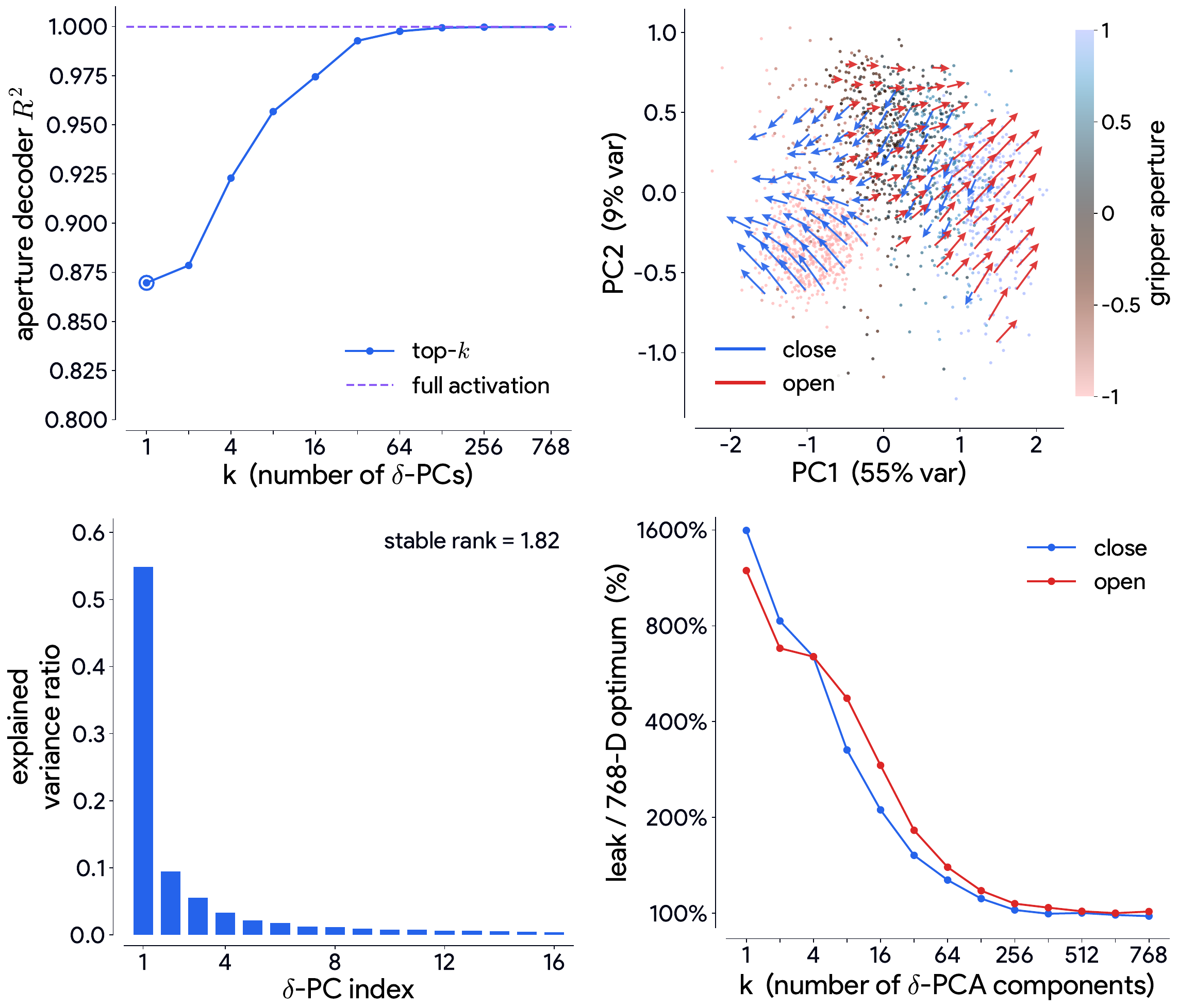}
    \caption{
    \textbf{Gripper control for action expert layer 9}
    (Top left) A linear probe decodes the commanded gripper aperture; one principal component of the pullback steering vectors decodes it nearly as well as the full activation.
    (Top right) Natural-rollout activations colored by gripper state; notably, the steering vectors from each cluster do not point toward the other, unlike layer 32 in Figure \ref{fig:gripper}.
    (Bottom left) One component captures ${\sim}55\%$ of steering-vector variance.
    (Bottom right) One component flips the gripper but leaks into the arm joints relative to using the full 768-dimensional steering vector; more components remove this excess leakage.
    }
    \label{fig:gripper_l9}
\end{figure}

\begin{figure}[tb]
    \centering
    \includegraphics[width=\linewidth]{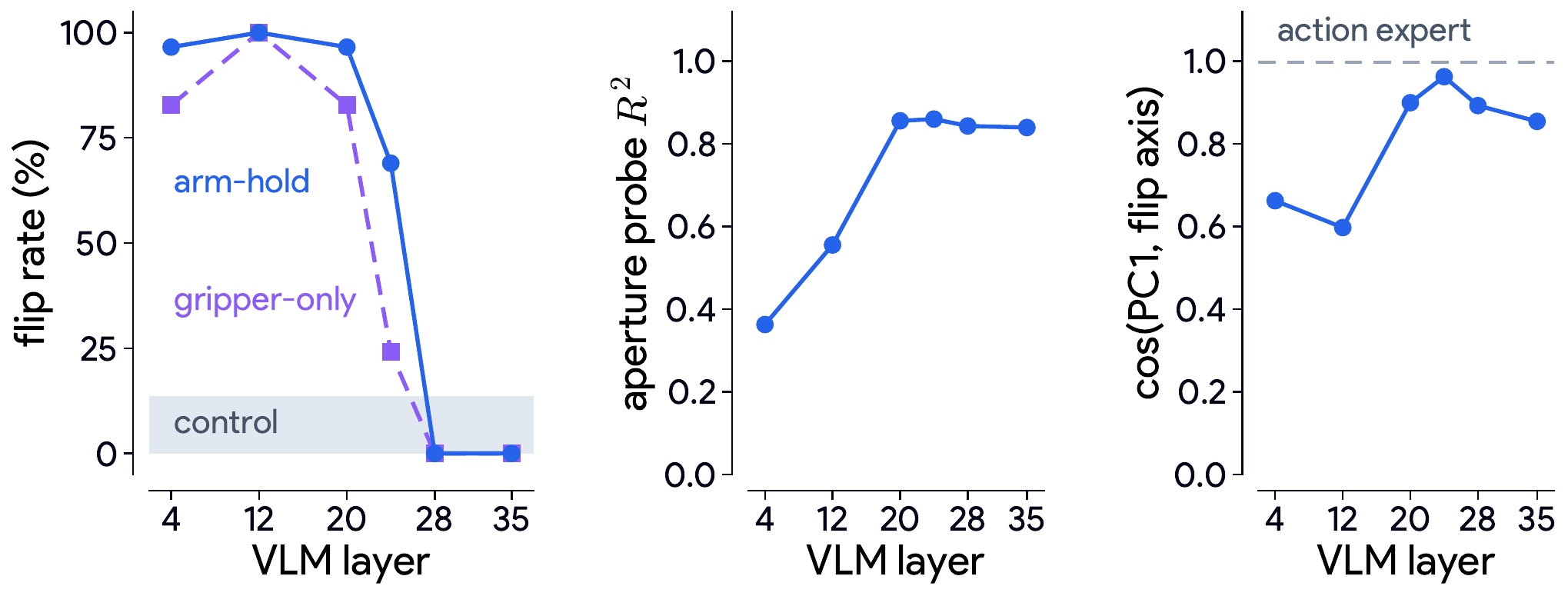}
    \caption{\textbf{In the VLM backbone, gripper steerability falls to the control floor by layer 28 although aperture readability stays high.}
    The gripper pullback of App.~\ref{app:gripper} with the edit $\delta$ added to every sequence token after VLM block $L$, at screen scale (32 edits per layer and objective, one query per landmark).
    (Left) Fraction of edits that flip the commanded gripper under the arm-hold objective of App.~\ref{app:gripper} and under the gripper-only objective (the arm-hold and smoothness terms dropped), counted on the 29 edits whose unedited chunk has the opposite gripper sign; the gray band spans the flip rates of a control edit (the mean of the action-expert catalog, App.~\ref{app:gripper}).
    (Middle) Held-out $R^2$ of a ridge probe from the pre-edit activation to the commanded aperture, best of the token pools at each layer; readability rises to 0.86 by layer 20.
    (Right) Absolute cosine between the first principal component of the arm-hold edits, a sample-limited estimate from the small per-layer catalog, and the close--open flip axis (dashed: the action-expert catalog of Figure~\ref{fig:gripper_l9}); at the shallowest flipping layers (4 and 12) the leading component is only partly aligned with the flip axis.
    }
    \label{fig:gripper_vlm_depth}
\end{figure}

\begin{figure}[tb]
    \centering
    \includegraphics[width=\linewidth]{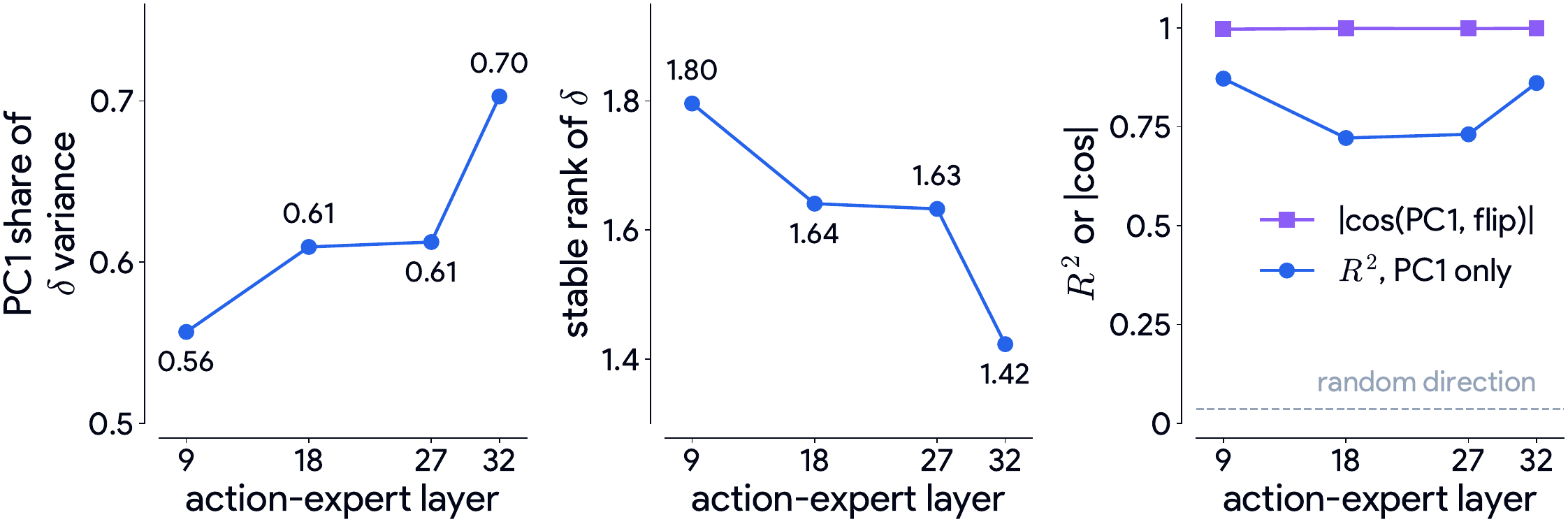}
    \caption{\textbf{The gripper steering vectors stay dominated by one component and aligned with the flip axis at every action-expert layer.}
    The pullback catalog of App.~\ref{app:gripper}, re-fit at layers 9, 18, 27, and 32. (Left) Share of steering-vector variance on the first principal component. (Middle) Stable rank of the steering vectors. (Right) Cosine between the first principal component and the flip axis (mean close edit minus mean open edit), and held-out $R^2$ of the aperture probe restricted to that component; dashed: a random direction.}
    \label{fig:gripper_depth}
\end{figure}

\paragraph{Gripper (App.~\ref{app:gripper}).}
Figure~\ref{fig:gripper_cumulative_var} plots the cumulative variance of the layer-32 steering vectors, of which the first component captures about 70\%.
Figure~\ref{fig:gripper_l9} repeats the analyses of Figure~\ref{fig:gripper} at action-expert layer 9, where one component again decodes and flips the gripper, but the steering vectors do not point from one gripper-state cluster toward the other.
Figure~\ref{fig:gripper_depth} refits the steering vectors at layers 9, 18, 27, and 32 and finds them dominated at every depth by one component aligned with the close--open flip axis.
Figure~\ref{fig:gripper_vlm_depth} applies the same pullback in the VLM backbone, where flip rates fall to the level of a control edit by layer 28, although the gripper command remains readable there.

\begin{figure}[tb]
    \centering
    \includegraphics[width=\linewidth]{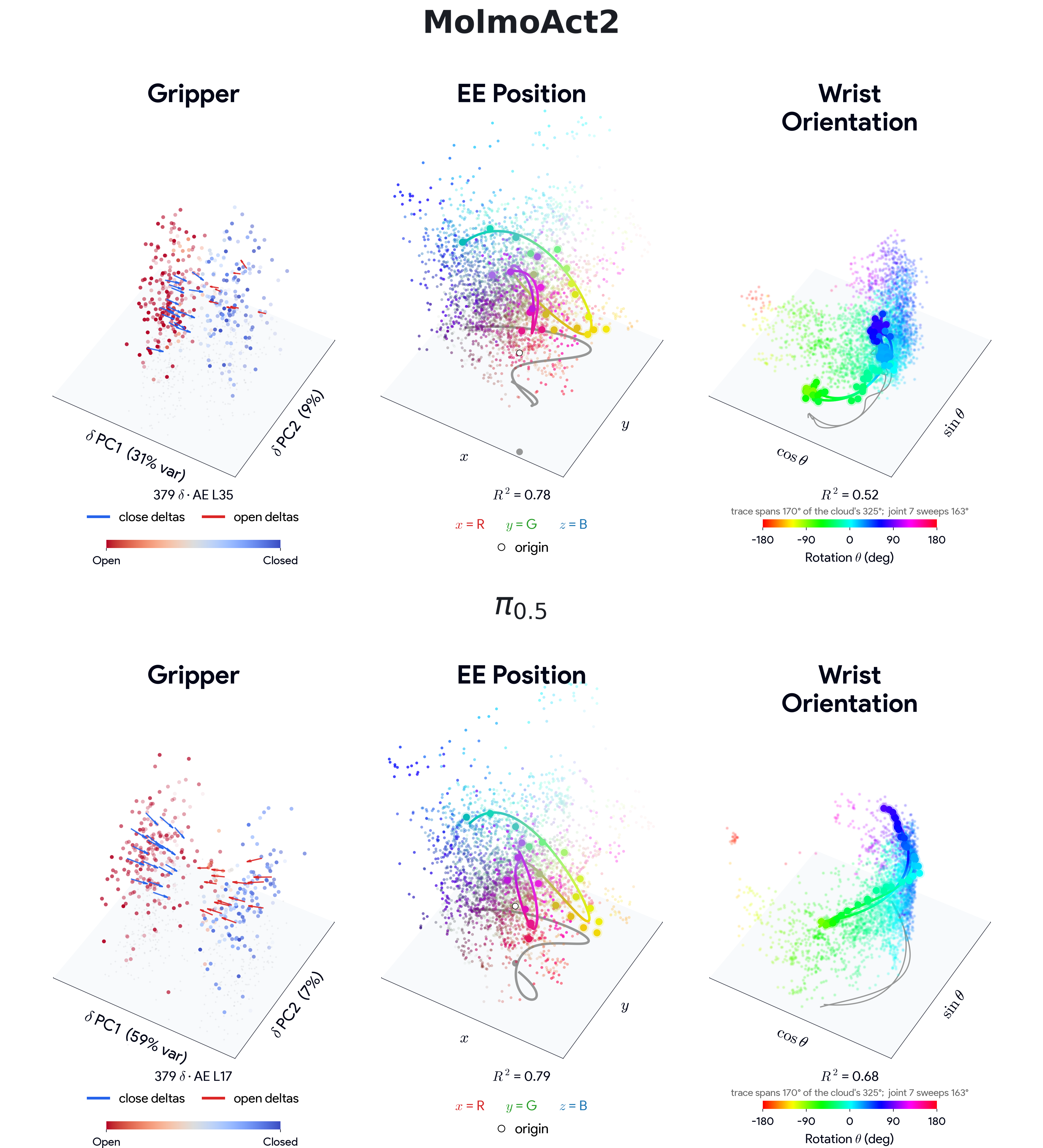}
    \caption{\textbf{DROID real world behavioral manifolds.} Here we show the behavioral manifolds realized by MolmoAct2 and $\pi_{0.5}$. The EE position and wrist orientation point clouds show the activations of all 3{,}321 queries, sampled from MolmoAct2 AE L16 of 36 and $\pi_{0.5}$ VLM L7 of 18, demonstrating the structures appear across multiple streams. We plot a natural episode trajectory in the EE position subplots and the wrist roll sweep of Figure~\ref{fig:wrist_real} in the Wrist Orientation subplots. In the Gripper plot we project the 379 landmark activations from each model's final action-expert block onto the principal components of the pullback steering vectors, with binned steering vectors as arrows.}

    \label{fig:droid_manifolds}
\end{figure}

\begin{figure}[tb]
    \centering
    \includegraphics[width=\linewidth]{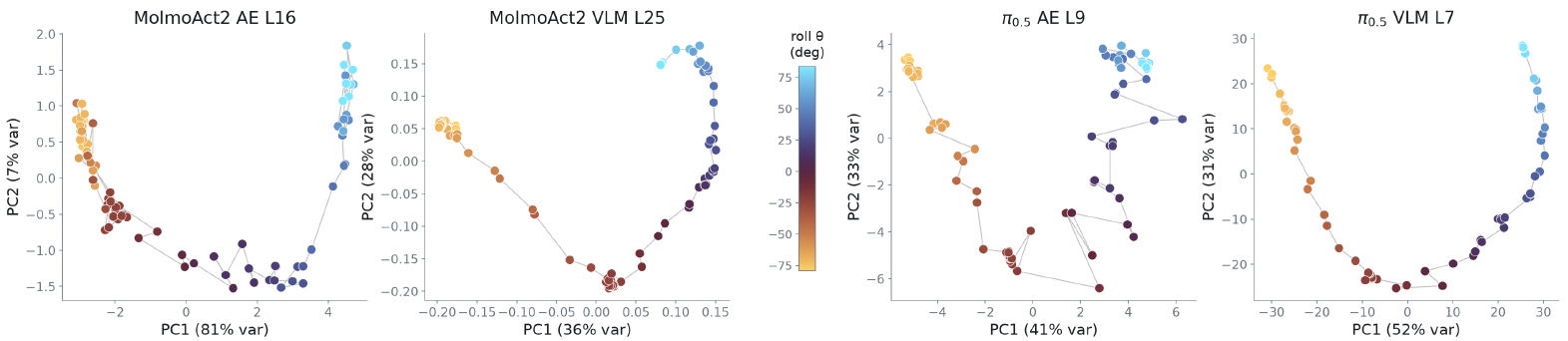}
    \caption{\textbf{Wrist orientation PCA across models and streams.} Following Fig.~\ref{fig:wrist}, we record a real-robot episode in which the wrist roll joint is swept from $-79^\circ$ to $+84^\circ$ while the other joints and the scene are held approximately fixed, and plot the top two principal components of the activations for MolmoAct2 and $\pi_{0.5}$ in both the VLM and the action expert (color: roll angle). As in MolmoBot, the sweep traces an open, semicircular arc. Because real-world data are noisier, the trajectories are considerably less smooth.}
    \label{fig:wrist_real}
\end{figure}

\begin{figure}[tb]
    \centering
    \includegraphics[width=\linewidth]{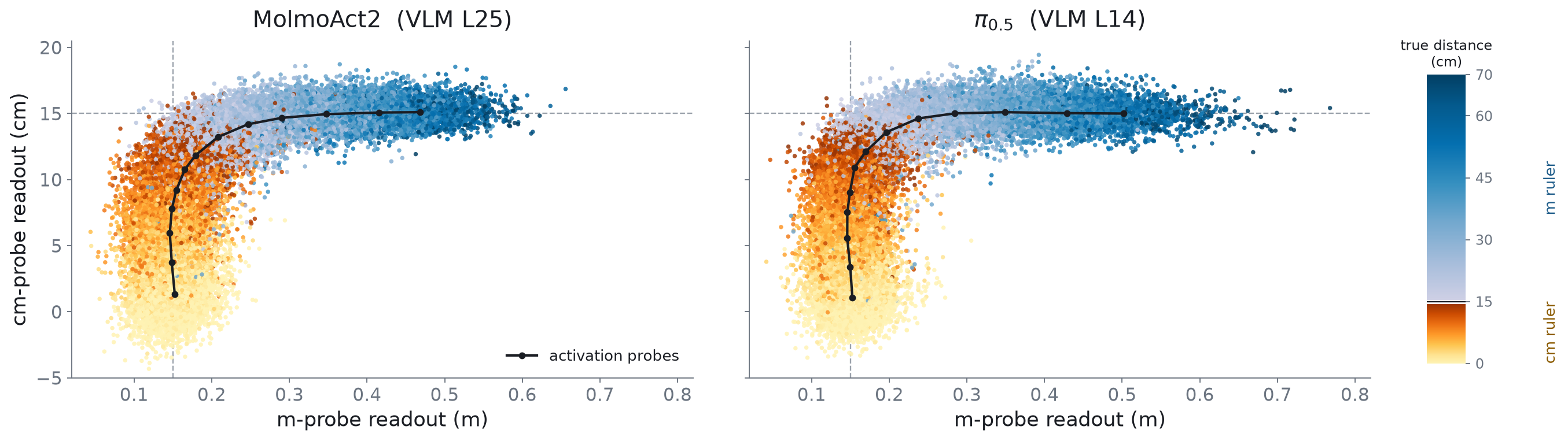}
    \caption{\textbf{MolmoAct2 and $\pi_{0.5}$ encode distance to target object in 2 orthogonal subspaces.} We find that similar to MolmoBot, as shown in Fig. \ref{fig:dist_target} the tested DROID model also encode distance to target object with a multi-scale ruler. We sample 26,675 timesteps from 1,998 pick and place episodes. The distance is how far the gripper is from its initial grasp location. Activations are sampled from MolmoAct2 VLM L25 of 36 and $\pi_{0.5}$ VLM L14 of 18.}
    \label{fig:droid_dist_target}
\end{figure}

\begin{figure}
    \centering
    \includegraphics[width=\linewidth]{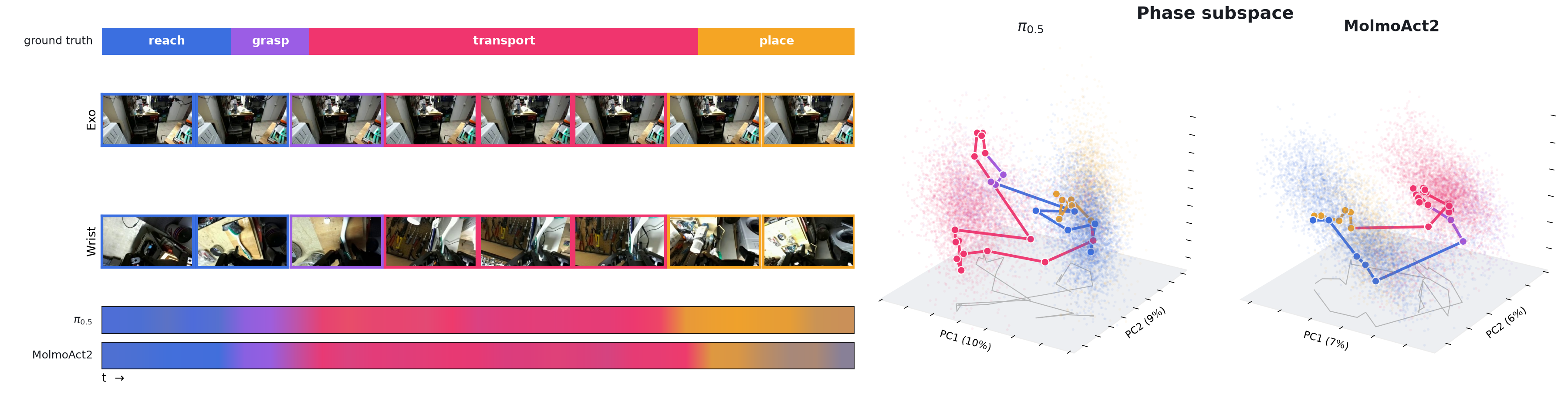}
    \caption{\textbf{Task phase subspace in $\pi_{0.5}$ and MolmoAct2.} (Left)
    Example trajectory for a simple pick-and-place task, ``Pick up the white and
    green brush from the table and hang it on the wall,'' along with the
    predicted task phases. The ground-truth bar is derived from the measured
    gripper aperture alone, using a simple heuristic: reach until the gripper
    starts to close, grasp while it closes, transport while it stays closed,
    place after it opens. A linear classifier is trained on the top 48 principal
    components of activations collected across 2{,}000 episodes. (Right) The
    recovered task subspace in both models, colored by phase. Activations are
    taken from the $\pi_{0.5}$ action expert at layer 9 of 18 and the MolmoAct2
    action expert at layer 22 of 36.}
    \label{fig:phase_droid}
\end{figure}

\begin{figure}
    \centering
    \includegraphics[width=1.0\linewidth]{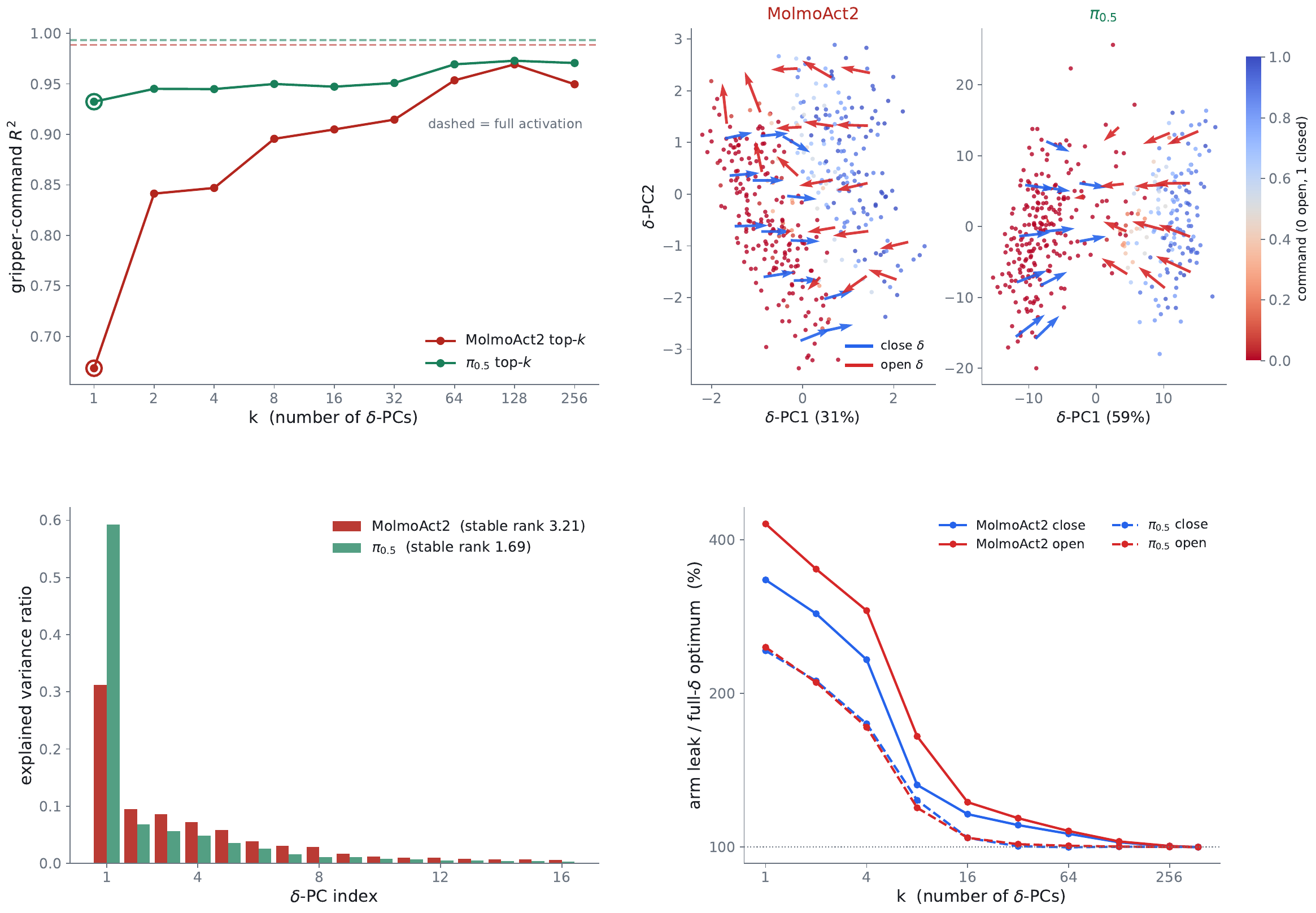}
    \caption{\textbf{Gripper steering vectors on real DROID episodes}. (Top left) A linear probe decodes each model's own gripper command from the activations projected onto the top-k principal components of the steering vectors. Dashed lines: each model's full-activation ceiling. (Top right) Activations colored by the commanded gripper (red open, blue closed); the steering vectors point from each cluster toward the other.  (Bottom left) One component captures 59 \% of $\pi_{0.5}$'s steering-vector variance (stable rank 1.69) but only 31\% of MolmoAct2's (stable rank 3.21).(Bottom right) One component flips the gripper (94–96\% of attempts, both models) but leaks into the arm joints at 2.4–4.3× the full-$\delta$ optimum. Activations are recorded and steered at the final action-expert block (MolmoAct2 layer 35 of 36, $\pi_{0.5}$ layer 17 of 18, counting from 0).}
    \label{fig:gripper_droid}
\end{figure}

\paragraph{Real robots (App.~\ref{app:real}).}
Figures~\ref{fig:droid_manifolds}--\ref{fig:gripper_droid} repeat the analyses on real-robot data for MolmoAct2 and $\pi_{0.5}$: DROID episodes, plus one recorded wrist-roll episode.
Figure~\ref{fig:droid_manifolds} shows the gripper, EE position, and wrist orientation structures in both models.
Figure~\ref{fig:wrist_real} shows that a real wrist-roll sweep traces an open, semicircular arc in both streams of both models (for the $\pi_{0.5}$ action expert, only with a fixed noise seed at an intermediate flow step).
Figure~\ref{fig:droid_dist_target} shows the two distance rulers: far points spread along the meter ruler and near points along the centimeter ruler.
Figure~\ref{fig:phase_droid} shows that the four task phases are decodable along a real episode, and that reach and transport occupy separate regions of the phase subspace in both models.
Figure~\ref{fig:gripper_droid} repeats the gripper pullback: one steering-vector component flips the gripper in both models, but it carries 59\% of the steering-vector variance in $\pi_{0.5}$ and 31\% in MolmoAct2.

\clearpage

\iffalse
\begin{figure}[h]
    \centering
    \includegraphics[width=\linewidth]{assets/cos_distance.pdf}
    \caption{\textbf{Near and far distance probes are nearly orthogonal.} Per boundary $c$, a probe pair is fit either on close and far distances separately (range-split) or on all distances with targets saturated at $c$ (round-to-$c$); lines show the cosine between the pair's weight vectors (fold mean $\pm$ s.d.). 
    Dashed: each pair's split-half ceiling --- the cosine expected if both probes measured the same axis. At the 15\,cm cutoff, cosines are $0.04$ and $0.07$ against ceilings of ${\sim}0.18$ and ${\sim}0.27$; the range-split cosine reaches its ceiling only at $c \gtrsim 40$,
    where the near fit degenerates to the global axis.}
    \label{fig:cos_distance}
\end{figure}
\fi

\clearpage

\clearpage
\begin{figure}[t]
    \centering
    \includegraphics[width=\linewidth]{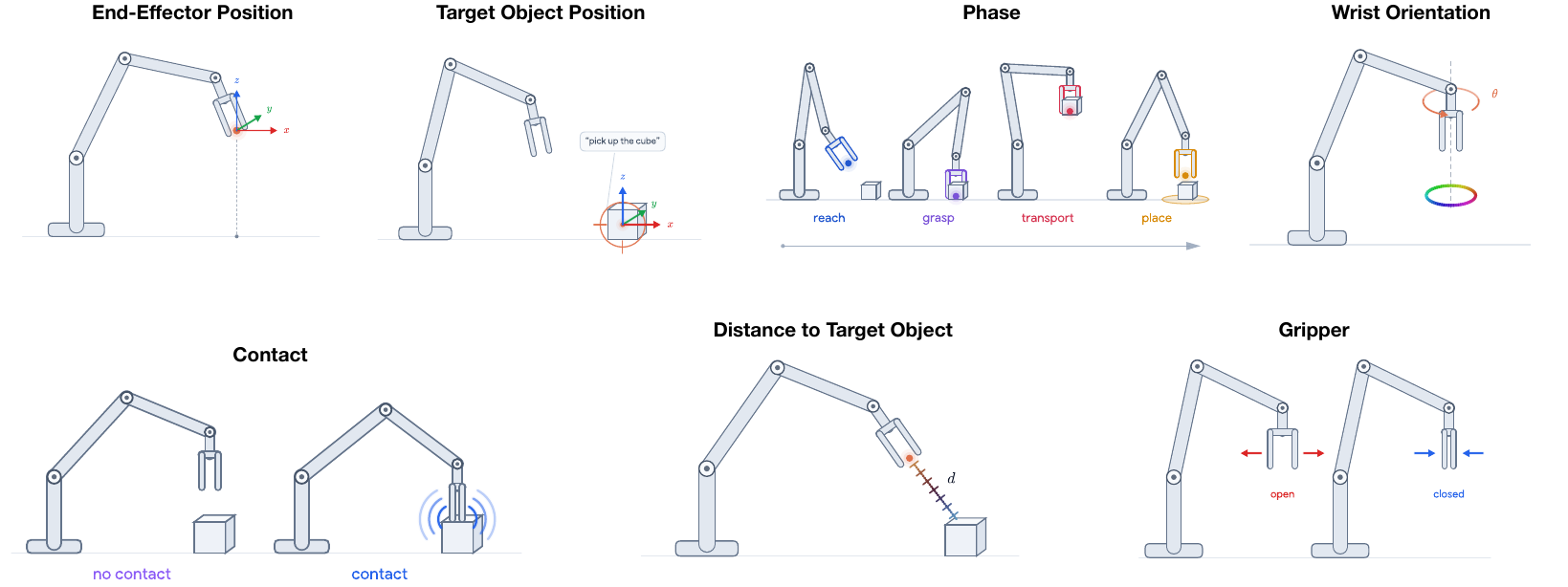}
    \caption{\textbf{The seven task variables.} An overview of each variable we study.}
    \label{fig:variables}
\end{figure}

%% file: experimental_details.tex
\section{Experimental Details}
\label{app:details}

\subsection{MolmoBot Architecture}
\label{app:architecture}

MolmoBot \citep{deshpande2026molmob0t} consists of two streams: a vision-language model (VLM) and an action expert. The VLM takes in the language instruction and images from two cameras: an exocentric camera providing a fixed third-person view of the workspace, and a wrist camera mounted on the end effector providing a first-person view that moves with the gripper. At each query after the first, the VLM receives the current frame and a frame from eight steps earlier from each camera; the first query of an episode, including every state recorded right after a teleport, sees only the current frames. The VLM receives no proprioceptive input, so any information it carries about the robot's state must be recovered from these observations.

The action expert takes in the robot's proprioceptive state together with noisy action tokens, which it denoises into an action chunk over 10 steps of flow matching. The two streams contain 36 layers each, and each layer of the action expert cross-attends to the VLM layer at the same depth, so visual and language information can enter the action computation at every level. Information flows in one direction only: the action expert reads from the VLM, but the VLM does not attend to the action expert. The action expert can therefore access the target object, which appears only in the camera images, solely through this cross-attention.

\subsection{Data}
\label{app:harvests}
Simulation experiments run in MolmoSpaces on ProcTHOR \citep{deitke2022} houses. All use the \emph{pick} task (``Pick up the X'') except phase, which uses \emph{pick-and-place} (``Pick up the X and place it in or on the Y''). Every probe is 5-fold cross-validated with houses held out. We collect four datasets:
\begin{enumerate}[nosep,leftmargin=*]
\item \emph{Natural pick rollouts} (1{,}008 episodes, 196 houses), with action-expert activations at every action-chunk query. Used for wrist orientation, the gripper pullback, and contact (for which we re-run 299 episodes in the same houses and with the same seeds to record VLM activations).
\item \emph{Object teleports} (906 scenes). The target is teleported to up to 16 placements per scene (Figure~\ref{fig:teleport_map}) and activations are recorded at the first four chunk boundaries, with an additional near-range set of placements within 35\,cm of the hand (68{,}174 states in total). Used for distance and direction.
\item \emph{Crossed arm${\times}$object teleports} (945 scenes). Five object placements are crossed with ten arm poses, recorded immediately after the teleport (q0) and after one executed chunk (q1). Used for both position variables.
\item \emph{Pick-and-place rollouts} (530 episodes, 181 houses, 63\% success). Used for phase.
\end{enumerate}
Position, distance, direction, and contact are read from the VLM residual stream at the token positions given below (a single final token, or the mean over a named set of tokens). Phase, wrist orientation, and the gripper are read from the action expert, averaged over the 16 action tokens at the final flow step unless noted. Layers are indexed from 0 in both streams.

\subsection{End-Effector and Target Object Position}
\label{app:position}
\paragraph{Data.}
In each scene we pick ten gripper targets at random in a box in front of the robot (20--60\,cm forward, up to 30\,cm to either side, 5--40\,cm above the object's resting position) and move the arm there with inverse kinematics. We keep a target only if the gripper gets within 1.5\,cm of it without the arm intersecting anything, and we drop arm--object pairs in which the gripper is within 12\,cm of the object. Each arm pose is paired with five object placements chosen the same way as in the object-teleport dataset. After each teleport the robot holds its new arm configuration, so the policy only sees frames taken after the move. Labels are the gripper and object positions read from the simulator after the teleport, in the robot's base frame.

\paragraph{Probes.}
For each variable we average the activations over a group of tokens and fit a ridge regression onto its $(x,y,z)$ position, with houses held out; both variables use the same rows and folds. We use the token group and layer that decode each variable best: for EE position, the non-image tokens before the instruction (the short labels that introduce each camera image, such as ``Image 1 View 1'', the image start and end markers, and the beginning-of-sequence token) at layer~20; for the target, the instruction tokens at layer~21. To count how many dimensions a readout uses, we restrict its predictions to their top one, two, or three principal directions (the rank curves) and report the singular values of the predictions. Table height differs between scenes but not within one, so part of the $z$ signal only identifies the scene; subtracting each scene's mean from the true and predicted positions removes it (within-scene $R^2$: 0.92 for EE, 0.57 for the target). The true $x$ and $y$ of each variable predict the other's at $R^2 \le 0.02$.

\paragraph{Equivariance.}
To test whether a subspace preserves spatial geometry, we take pairs of states from the same scene that differ in one variable only: two arm poses with the object fixed, or two object placements with the arm fixed (the object moves only horizontally, so its displacement is 2-D). On training scenes we fit a map from the world displacement to the resulting change in activation within the rank-3 subspace, and on held-out scenes we report the median relative error of its predictions, once for any linear map and once for a rotation with uniform scaling (Procrustes). Shuffling which world displacement goes with which activation change gives an error of 1.00.

\paragraph{Curvature.}
We move the hand along four straight lines (forward, sideways, up, and toward the object; 47 scenes) and the object along straight lines with the arm at its home pose (43 scenes), up to 24 points per line, and project the activations onto the subspace fit without those scenes. Curvature is the median angle between consecutive ${\sim}7$\,cm segments of the projected path, binned by hand--object distance and compared with the same activations projected onto random 3-D subspaces (median of eight). The paths in Figure~\ref{fig:position} were traced separately in one of these held-out scenes.

\paragraph{Action expert (Figure~\ref{fig:position_ae}).}
Applied to the action expert's activations, the same pipeline decodes EE position best at layer~12 ($R^2 = 0.998$, 1.1\,cm error). There, a general linear map predicts the displacements almost exactly (error 0.04) but a rotation with uniform scaling does not (0.29), so the subspace is a stretched rather than a rigid copy of world coordinates.

\subsection{Distance to Target Object}
\label{app:distance}
\paragraph{Data.}
We use the object-teleport dataset, keeping the 67{,}778 states in which the object is within 70\,cm of the gripper (905 scenes; 11{,}534 states below 15\,cm). Distance $d$ is the straight-line distance between the gripper and the object, read from the simulator. Activations are the VLM's final token at layer~21; the action expert's last layer (35) gives the same main results (Figure~\ref{fig:distance_ae35}). The erasure test, the left panel of Figure~\ref{fig:distance_link_race}, and the EE control below use only the first round of teleports, recorded before the extra near-range placements were added (52{,}210 states; 4{,}422 below 15\,cm).

\paragraph{Erasure (Figure~\ref{fig:dist_target}a).}
We split states into near (below 15\,cm) and far (15--70\,cm). Within each cross-validation fold, we fit a linear probe for distance on one range, remove its direction from all activations, and repeat, fitting each new probe on the already-cleaned activations and making its direction orthogonal to those already removed, up to 32 times. After each removal we refit a probe on each range and report its held-out $R^2$, computed against the variance within that range.

\paragraph{Rulers and cutoff sweep.}
The meter ruler is a ridge probe fit on all states to distance with every value below 15\,cm set to 15\,cm; the centimeter ruler is the mirror image, with every value above 15\,cm set to 15\,cm. Each is scored against the true distance within its own range. To judge whether two probes point in different directions, we compare the cosine between them with the cosine expected if both measured the same direction: for each probe, the cosine between two copies fit on disjoint halves of the houses (median of three random splits), taking the lower of the two probes' values. The cutoff sweep repeats this with the cutoff moved from 3 to 56\,cm, fitting the pair either only on the states on each side of the cutoff (range-split) or on all states with values clipped at the cutoff (round-to-$c$), and compares both with a single probe fit on the whole range and then rescaled and shifted separately on each side of the cutoff.

\paragraph{Controls.}
The near axis (the direction of the near-range probe) is not a disguised copy of EE position: EE position predicts nothing about distance among near states ($R^2 \approx 0$), and less than 3\% of the near axis falls within the subspaces that encode EE and target position. It does not detect contact either: the centimeter ruler's error is the same on states in which the object touches nothing but its supporting surface (1.46 vs.\ 1.45\,cm). For the single-direction control, we build synthetic activations in which distance is written along one random direction, plus noise and a per-scene offset, and run the same probes on them.

\subsection{Direction to Target Object}
\label{app:direction}
We use the same data and layer as for distance. The target is the unit bearing $\bm{u}$ from EE to object in the base frame, decoded by ridge regression onto its three components. Error is the angle between decoded and true bearing, compared to a training-mean baseline ($33.1^\circ$) and a scene-mean oracle ($19.0^\circ$). Since the bearing is a nonlinear function of the two positions, we check that the model does not simply recompute it from them: the residual of $\bm{u}$ after regressing out $[\bm{p}_{\mathrm{EE}}, \bm{p}_{\mathrm{obj}}, d]$ remains decodable ($R^2$ of 0.3--0.5 per component), and the residual probe has little energy in the position and distance subspaces (0.01--0.04). In Figure~\ref{fig:direction_vlm}, decoded bearings are averaged in 6\,cm cells of a target-centered frame, and the cutoff sweep follows the distance protocol.

\subsection{Contact}
\label{app:contact}
\paragraph{Data and probe.}
We record VLM activations by re-running the policy in the same house and with the same seed as each of 299 natural pick episodes. The re-runs are not copies of the originals: in 259 of them a different target object is sampled, and bf16 inference makes even repeated runs differ, so each state is labeled from the re-run itself: contact means some part of the gripper touches the object with nonzero contact force. We average the layer-24 activations over the text tokens of the prompt (the instruction and its chat-template words) and fit a linear discriminant (LDA) for contact, with houses held out. It reaches 91.3\% held-out accuracy against a 65\% majority baseline (AUC 0.954); the final token alone gives the same accuracy (AUC 0.957). Figure~\ref{fig:contact}a is a PCA of the same pooled activations. For Figure~\ref{fig:contact}b and the tests below, the probe is never trained on intervention states: each test pose is scored by the probe refit on natural states with that pose's house held out, so no tested scene is in its training data.

\paragraph{Off-schedule interventions.}
To make contact start or end at moments when it would not naturally change, we stop 123 natural pick episodes in the same houses (100 of them among the 299) at up to four poses of each kind, hold the arm fixed, and teleport the object in 1\,cm steps along the line between the gripper and the object, continuing three steps past the change: toward the gripper at poses where the policy commands it open, until the object first touches it (early touch), or away from the gripper at poses where it touches the object, usually in a closed grasp, until contact breaks (induced slip). After each step, both cameras are re-rendered from the arm's current pose. We repeat each test with the gripper forced to the opposite state: closed during early touches and open during slips. Forcing the gripper closed during early touches does not lower detection; forced-open slips (24 poses, as opening the gripper at a held grasp usually releases the object, which leaves few true slips) still score contact higher at most poses (75\%), but they are too few to compare with natural slips.

\subsection{Phase}
\label{app:phase}
\paragraph{Labels.}
Each step is labeled from the simulator state: the gripper command, the object's height, and whether the object rests on the receptacle. \textsc{Reach} lasts until four steps before the first command to close the gripper. \textsc{Grasp} runs from there until transport begins, so failed grasps and retries count as grasp. \textsc{Transport} begins once the object, held after the final close, has risen 3\,mm. \textsc{Place} begins where the object starts its last uninterrupted descent (at least 1\,cm) before being released onto the receptacle, or at the release if there is no such descent, and runs to the end of the episode.

\paragraph{Decoding.}
We fit a four-class linear discriminant (LDA) on the successful episodes, with houses held out. Balanced accuracy (the mean of the four per-phase accuracies; chance 0.25) is 0.92 for the action expert at layer~24, 0.86 for the VLM's image tokens of the two current camera frames at layer~31, and 0.87 for the VLM's non-image tokens at layer~29. The VLM's 3-D view (Figure~\ref{fig:phase-vlm}) shows the top three LDA directions rather than principal components, because the VLM's first principal component (78\% of variance) tracks the scene rather than the phase; the action expert's view in Figure~\ref{fig:phase}b uses principal components.

\paragraph{Freezing.}
The policy sees two frames per query, taken eight steps apart. At landmark moments in each phase (for place, the last query before the release), taken from re-simulations of successful episodes, we query the policy twice at the same moment: once with its normal two frames (natural), and once with the current frame shown twice, so that the world appears to have stopped (frozen; ``2-query pause'' in Figure~\ref{fig:phase-freeze}). Re-simulations that end in failure are kept. The decoded phase of both queries (Figures~\ref{fig:phase}c, \ref{fig:phase-vlm}, and \ref{fig:phase-frozen-matrices}) comes from the LDA of the previous paragraph, using the fold that held out each landmark's house (797 landmarks from 122 episodes). The gripper effect comes from all episodes of the same capture (819 landmarks from 126 episodes): the fraction of the eight executed steps that command a closed gripper, frozen minus natural (Figure~\ref{fig:phase}d; per step in Figure~\ref{fig:phase-freeze}).

\subsection{Wrist Orientation}
\label{app:wrist}
\paragraph{Roll sweep (Figures~\ref{fig:headline} and \ref{fig:wrist}).}
Starting from one arm pose, we set the wrist joint (Franka joint~7) to 38 evenly spaced angles across its range plus its starting angle, keep all other joints fixed, and re-render both cameras at each angle. We record the action expert at layer~9, using the first action token at the final flow step, and plot the top two principal components of the 39 activations. For Figure~\ref{fig:headline}, we run the same sweep in 64 houses, subtract each house's mean activation, and fit one PCA to all of them.

\paragraph{Readout and precession (Figures~\ref{fig:wrist_rank} and \ref{fig:wrist_circle}).}
The natural-rollout readout uses the first action token at the final flow step. In the precession sweep, each joint is kept within the range of the policy's training actions (1st to 99th percentile) and the hand within 5\,cm of its starting position; in each direction, the orbit stops when the hand can no longer be held there.

\paragraph{Steering (Figure~\ref{fig:wrist}b--d).}
The autoencoder is a pair of small MLPs (768--512--256--32 and back), trained on layer-9 activations averaged over the action tokens, from controlled wrist-rotation sweeps with a third of the houses held out. Three pairs of its 32 code dimensions are each constrained to a circle, one per rotation axis, and the other 26 are unconstrained; we rotate only the pair for rotation about the vertical axis, where one degree on the circle corresponds to one degree of physical rotation on held-out sweeps. To rotate by $\theta$, we encode the current activation, rotate its point on that circle by $\theta$, and decode. The edit is the decoded rotated point minus the decoded unrotated one, so a zero-degree rotation gives no edit; it is added to all action tokens at layer~9 at the final flow step and is $19\% \pm 7\%$ of the activation's norm. One-shot edits of $\pm 20^\circ$ and $\pm 40^\circ$ are applied at a single query in the middle of a rollout, in eight scenes held out from training the autoencoder, and the response is read from the decoded 8-step action chunk; one scene in which the wrist rotates strongly on its own is excluded, leaving 28 conditions. Held edits are recomputed from the live activation at every query over 32-step rollouts in six held-out scenes (five of these seven plus the excluded one), and the rotation is measured relative to the unedited rollout of the same scene.

\subsection{Gripper Pullback}
\label{app:gripper}
\paragraph{Landmarks.}
In 304 grasping episodes (170 houses), we take two landmarks relative to the moment the grasp becomes stable (contact held for three consecutive queries, with a lift of at least 2\,cm when one occurs): a few queries before it (typically 3--9, gripper open) and a few after (gripper closed and holding the object). At each landmark we pick up to three evenly spaced queries whose executed gripper command is opposite the target, giving 1{,}666 edits. For 304 of them the unedited chunk already has the target's sign; flip rates exclude these.

\paragraph{Objective.}
We fit an additive edit $\bm{\delta} \in \mathbb{R}^{768}$ to all 16 action tokens at the output of action-expert block $L$, at the final flow step only, with the model's weights fixed. Let $a(\bm{\delta}) \in \mathbb{R}^{H\times 8}$ be the decoded chunk ($H{=}16$), with seven arm channels $a^{\mathrm{arm}}$ and a gripper channel $g$. We minimize
\begin{equation*}
\small
\begin{aligned}
\mathcal{L}(\bm{\delta}) \;=\;
   & \tfrac{w_g}{H}\textstyle\sum_{t}\bigl(g_t(\bm{\delta}) - g^\ast\bigr)^2
   \;+\; \tfrac{w_a}{7H}\textstyle\sum_{t,j}\bigl(a^{\mathrm{arm}}_{t,j}(\bm{\delta}) - \bar{a}_{j}\bigr)^2 \\
   &+\; \tfrac{w_s}{7(H-1)}\textstyle\sum_{t,j}\bigl(a^{\mathrm{arm}}_{t+1,j}(\bm{\delta}) - a^{\mathrm{arm}}_{t,j}(\bm{\delta})\bigr)^2
   \;+\; \lambda \lVert\bm{\delta}\rVert_2^2,
\end{aligned}
\end{equation*}
where $g^\ast \in \{+1,-1\}$ is the target gripper command (close, open) and $\bar{a}$ is the current joint configuration; the terms flip the gripper, hold the arm still, and keep the chunk smooth. We use $(w_g, w_a, w_s, \lambda) = (1, 300, 3000, 10^{-4})$, with $w_a$ and $w_s$ chosen by grid search at $w_g = 1$, and run Adam (learning rate 0.1) for 1{,}000 steps, keeping the lowest-loss edit. An edit succeeds if the gripper command, averaged over the chunk, has the sign of $g^\ast$; all measurements use the decoded chunk, which is not executed.

\paragraph{Low-rank edits and leakage.}
For the leakage curves we refit the same objective with the edit restricted to $\bm{\mu} + V\bm{w}$, where $\bm{\mu}$ and $V$ are the mean and top $k$ principal components of the edits from all other houses, at one query per landmark per episode. Leakage is the root-mean-square gap between the arm commands and the current joint angles over the chunk, reported as the median over queries relative to the median for the unrestricted 768-dimensional edits.

\paragraph{Probe and steering field.}
The aperture probe is a ridge regression, with houses held out, from each state's pre-edit activation, projected onto the top $k$ edit components, to the gripper command averaged over its decoded chunk. The field panel places pre-edit states on the top two components and averages the edits in each grid cell, separately for closing and opening edits.

\subsection{Real-Robot Analyses}
\label{app:real}
\paragraph{Data.}
We use DROID \citep{khazatsky2024droid}, giving each model one frame per camera at each query (no earlier frame). The first 100 episodes, queried every eighth step (3{,}321 queries), are used for Figures~\ref{fig:real_manifolds}, \ref{fig:droid_manifolds}, and \ref{fig:gripper_droid}; the wrist traces in the first two instead show the roll-sweep episode described below. A further 2{,}000 pick-and-place episodes, drawn at random from those with a single grasp-and-release cycle and at least six recorded queries after the release, are used for Figures~\ref{fig:droid_dist_target} and \ref{fig:phase_droid}. Action-expert activations are averaged over the action tokens at the final flow step. VLM activations are averaged over the image tokens, with each token first scaled to unit length for MolmoAct2 and MolmoBot; the MolmoAct2 distance rulers instead average the non-image tokens. Reported scores are 5-fold cross-validated with episodes held out. \emph{Closure onset} is the first query at which the measured gripper position leaves zero, ignoring blips that stay below 30\% of its holding value.

\paragraph{EE position and wrist orientation.}
Labels are computed from the recorded joint angles with the robot's kinematic model; the wrist angle is the gripper's rotation about the vertical axis (the last angle of its $xyz$ Euler decomposition). Probes read action-expert layer~16 of MolmoAct2 and MolmoBot and VLM layer~7 of $\pi_{0.5}$. For the roll sweep (Figure~\ref{fig:wrist_real}), an operator rotated joint~7 from $+69^\circ$ to $-79^\circ$ and back to $+84^\circ$ in one episode while the other joints moved by up to $15^\circ$; Figure~\ref{fig:wrist_real} colors points by this joint angle. The $\pi_{0.5}$ action-expert panel uses one fixed noise seed for all queries, the third of the ten flow steps, and the first action token (with a new seed per query at the final step, no arc is visible).

\paragraph{Distance to grasp.}
We use the queries from the start of each episode up to closure onset, keeping pick-and-place episodes with at least three such queries, and define $d$ as the gripper's distance from where it is at closure onset. The rulers are fit as in App.~\ref{app:distance}, on VLM layer~25 of MolmoAct2 and MolmoBot and layer~14 of $\pi_{0.5}$.

\paragraph{Phase.}
Labels come from the measured gripper position: reach until closure onset, grasp until the gripper reaches 90\% of its holding value, transport while it stays there, and place for the six queries (3.2\,s) after that; later queries are not scored. The bars in Figure~\ref{fig:phase_droid} come from four one-vs-rest linear discriminants, one per phase, each fit on all other episodes.

\paragraph{Gripper pullback.}
We follow App.~\ref{app:gripper}, adding $\bm{\delta}$ after the final action-expert block, with 15 action tokens for MolmoAct2 and $\pi_{0.5}$ (whose activations are 1{,}024-dimensional) and a gripper weight of 16 for MolmoAct2 and 4 for $\pi_{0.5}$ and MolmoBot. MolmoAct2's gripper command lies in $[0,1]$, so its targets are 0 and 1 and a flip means crossing 0.5; $\pi_{0.5}$ outputs joint velocities, so its arm target is zero velocity. All models share 379 landmarks: queries 3--9 before closure onset (gripper open) or 0--9 after (closing), at most two of each per episode. Rank-$k$ edits project each $\bm{\delta}$ onto the mean and top $k$ principal components of all 379 edits instead of refitting.

\paragraph{Cross-model similarity.}
MolmoBot is run on the same DROID queries. For each variable, we project each model's standardized activations onto that model's own ridge-probe directions for the variable, out of fold (aperture on the 379 landmark states; $(\cos\theta, \sin\theta)$ and $(x, y, z)$ on the 3{,}321 queries; the two rulers on the reach queries of the first 100 episodes). Similarity is one minus the Procrustes disparity between two models' coordinates on identical rows, after centering, scaling to unit norm, and the best rotation, reflection, and uniform scaling; the null repeats this with the target shuffled across rows. Every coordinate is fit to the same labels in each model, so part of the agreement is built in.